\documentclass[letterpaper]{article} 
\usepackage{AAAI_style/aaai23}  
\usepackage{times}  
\usepackage{helvet}  
\usepackage{courier}  
\usepackage[hyphens]{url}  
\usepackage{graphicx} 
\usepackage{natbib}  
\usepackage{caption} 
\usepackage{algorithm}
\usepackage{algpseudocode}

\usepackage{newfloat}
\usepackage{listings}
\DeclareCaptionStyle{ruled}{labelfont=normalfont,labelsep=colon,strut=off} 
\floatstyle{ruled}
\newfloat{listing}{tb}{lst}{}
\floatname{listing}{Listing}
\usepackage[dvipsnames]{xcolor}

\usepackage{caption}
\usepackage{subcaption}

\usepackage{enumitem,kantlipsum}

\usepackage[utf8]{inputenc} 
\usepackage[T1]{fontenc}    
\usepackage{hyperref}       
\hypersetup{
pdftitle={AI for Global Climate Cooperation: Modeling Global Climate Negotiations, Agreements, and Long-Term Cooperation in RICE-N},
pdfsubject={
Comprehensive global cooperation is essential to limit global temperature increases while continuing economic development, e.g., reducing severe inequality or achieving long-term economic growth.
Achieving long-term cooperation on climate change mitigation with n strategic agents poses a complex game-theoretic problem.
For example, agents may negotiate and reach climate agreements, but there is no central authority to enforce adherence to those agreements.
Hence, it is critical to design negotiation and agreement frameworks that foster cooperation, allow all agents to meet their individual policy objectives, and incentivize long-term adherence.
This is an interdisciplinary challenge that calls for collaboration between researchers in machine learning, economics, climate science, law, policy, ethics, and other fields.
In particular, we argue that machine learning is a critical tool to address the complexity of this domain.
To facilitate this research, here we introduce RICE-N, a multi-region integrated assessment model that simulates the global climate and economy, and which can be used to design and evaluate the strategic outcomes for different negotiation and agreement frameworks.
We also describe how to use multi-agent reinforcement learning to train rational agents using RICE-N. 
This framework underpinsAI for Global Climate Cooperation, a working group collaboration and competition on climate negotiation and agreement design.
Here, we invite the scientific community to design and evaluate their solutions using RICE-N, machine learning, economic intuition, and other domain knowledge. 
More information can be found on www.ai4climatecoop.org.
},
pdfauthor={
Tianyu Zhang,
Andrew Williams,
Soham Phade,
Sunil Srinivasa,
Yang Zhang,
Prateek Gupta,
Yoshua Bengio, 
Stephan Zheng
},
pdfkeywords={Climate change, climate negotiation, climate agreements, machine learning, game theory, integrated assessment models, multi-agent reinforcement learning}
}

\usepackage{booktabs}       
\usepackage{amsfonts}       
\usepackage{nicefrac}       
\usepackage{microtype}      
\usepackage{lipsum}
\usepackage{graphicx}       
\graphicspath{{media/}}     

\usepackage{amsmath}
\usepackage{amssymb}
\usepackage{amsthm}
\usepackage{bm}

\newcommand\blfootnote[1]{%
  \begingroup
  \renewcommand\thefootnote{}\footnote{#1}%
  \addtocounter{footnote}{-1}%
  \endgroup
}

\title{
AI for Global Climate Cooperation:
Modeling Global Climate Negotiations, Agreements, and Long-Term Cooperation in \SimulationName{}
}

\author{
Tianyu Zhang\textsuperscript{\rm 1, 2},
Andrew Williams\textsuperscript{\rm 1, 2},
Soham Phade\textsuperscript{\rm 6},
Sunil Srinivasa\textsuperscript{\rm 6}\thanks{Work done while at Salesforce.},
Yang Zhang\textsuperscript{\rm 2},\\
Prateek Gupta\textsuperscript{\rm 2, 3, 4},
Yoshua Bengio\textsuperscript{\rm 1, 2, 5}, 
Stephan Zheng\textsuperscript{\rm 6}%
}
\affiliations {
    \textsuperscript{\rm 1} Universit\'e de Montr\'eal, \textsuperscript{\rm 2} Mila, \textsuperscript{\rm 3} University of Oxford, \\ 
    \textsuperscript{\rm 4} The Alan Turing Institute, \textsuperscript{\rm 5} CIFAR, \textsuperscript{\rm 6} Salesforce Research
}

\newcommand{\comment}[1]{}

\def\eqref#1{equation~\ref{#1}}

\DeclareMathAlphabet{\mathsfit}{\encodingdefault}{\sfdefault}{m}{sl}
\SetMathAlphabet{\mathsfit}{bold}{\encodingdefault}{\sfdefault}{bx}{n}

\newcommand{\calA}{\mathcal{A}}

\newcommand{\calC}{\mathcal{C}}

\newcommand{\calE}{\mathcal{E}}

\newcommand{\calI}{\mathcal{I}}

\newcommand{\E}{\mathbb{E}}

\DeclareMathOperator*{\argmax}{arg\,max}

\newcommand{\eq}[1]{\begin{align}#1\end{align}}

\newcommand{\fr}[2]{\frac{#1}{#2}}
\newcommand{\brck}[1]{\left(#1\right)}

\newcommand{\brcksq}[1]{\left[#1\right]}
\newcommand{\brckcur}[1]{\left\{#1\right\}}

\def\shownotes{1}  
\ifnum\shownotes=1
\newcommand{\authnote}[2]{{\scriptsize $\ll$\textsf{#1 notes: #2}$\gg$}}
\else
\newcommand{\authnote}[2]{}
\fi
\ifnum\shownotes=1

\else

\fi

\newcommand{\srp}[1]{{\color{brown}\authnote{Soham}{#1}}}

\def\SimulationName{RICE-N}

\def\climateindex{\calC}
\def\economicindex{\calE}
\def\idstep{t}
\def\idxi{i}
\def\IndexSetI{\calI}
\def\idxj{j}
\def\idxk{k}
\def\numEpisodeSteps{H}
\def\numAgents{n}
\def\numRegions{\numAgents}
\def\numStepPeriod{\Delta}
\def\xcapital{K}
\def\xcapitaldecay{\Phi_{K}}
\def\xcapitalelasticity{\gamma}
\def\xcarbonintensity{\sigma}
\def\xcarbonmass{M}

\def\xconsumption{C}
\def\xemission{E}

\def\xforcing{F}
\def\xgoodstraded{x}
\def\xproduction{Y}
\def\xgrossoutput{Q}
\def\ximportbid{b}
\def\ximportbidmatrix{B}
\def\xexportlimit{p^{x}}
\def\xinvest{I}
\def\xmaxexportlimit{p^{x}}
\def\xmaxexport{x^\text{max}}
\def\xmitigation{\mu}
\def\xmitigation{\mu} 
\def\xnetoutput{Q}

\def\xpopulation{L}
\def\xsavings{s}
\def\xtariff{\tau}
\def\xtechnology{A}

\def\xtemperature{T}
\def\xtradebalance{D}
\def\xdebtratio{d}

\def\xutility{U}

\newcommand{\policy}{\pi}

\def\ob{o}

\def\ac{a}
\def\ActionSpace{\calA}

\def\rew{r}

\def\df{\gamma}

\def\eplen{T}

\newcommand{\degreeC}[1]{#1^\circ \mathrm{C}}

\begin{document}
\maketitle

\begin{abstract}
Comprehensive global cooperation is essential to limit global temperature increases while continuing economic development, e.g., reducing severe inequality or achieving long-term economic growth.
%
%
Achieving long-term cooperation on climate change mitigation with $\numRegions$ strategic agents poses a complex game-theoretic problem.
For example, agents may negotiate and reach climate agreements, but there is no central authority to enforce adherence to those agreements.
Hence, it is critical to design negotiation and agreement frameworks that foster cooperation, allow all agents to meet their individual policy objectives, and incentivize long-term adherence.
This is an interdisciplinary challenge that calls for collaboration between researchers in machine learning, economics, climate science, law, policy, ethics, and other fields.
In particular, we argue that machine learning is a critical tool to address the complexity of this domain.
To facilitate this research, here we introduce \SimulationName{}, a multi-region \emph{integrated assessment model} that simulates the global climate and economy, and which can be used to design and evaluate the strategic outcomes for different negotiation and agreement frameworks.
We also describe how to use multi-agent reinforcement learning to train rational agents using \SimulationName{}. 
This framework underpins \emph{AI for Global Climate Cooperation}, a working group collaboration and competition on climate negotiation and agreement design.
Here, we invite the scientific community to design and evaluate their solutions using \SimulationName{}, machine learning, economic intuition, and other domain knowledge. 
More information can be found on \url{www.ai4climatecoop.org}.
\end{abstract}




\section{Introduction}
%
The latest IPCC report~\cite{portner2022climate} warns that it is ``now or never'' to stave off a climate disaster.
Ecosystems are drastically changing: the Amazon rainforest is receding~\cite{lovejoy2018amazon} and polar ice sheets are melting~\cite{boers2021critical, deconto2021paris}.
Extreme weather events are clear warning signs too, such as the recent uptick in coastal flooding and forest fires. 
These developments are increasingly being attributed to climate change and driving towards a system-wide tipping point.

\blfootnote{
\textbf{For more information, see \url{www.ai4climatecoop.org}}.
}
\blfootnote{
E-mail: \emph{climate-cooperation-competition@googlegroups.com}.
}
\blfootnote{
Open-source code available at:
\url{https://github.com/mila-iqia/climate-cooperation-competition}.
}
\blfootnote{
Disclaimer: This project is for research purposes only. We do not intend to make normative or value statements about different social welfare objectives or policies.
}

Climate change is a global phenomenon, and it affects all.
In response, private and public financing have driven technological innovation (e.g., in renewable energy) and community campaigns to drive systemic change. 
However, climate change mitigation investments vary in size and type across nations and global regions, due to various social and economic factors.
For example, developing nations may need to focus first on the basic needs of their citizens, while developed nations have more funding and opportunities to prepare for the adverse impacts of climate change. 
Hence, climate change presents a classic case of ``tragedy of the commons'', where individual agents who pursue their own self-interest may lead to a destructive outcome for everyone.

As such, achieving and maintaining global cooperation is essential to achieve the Paris Agreement's long-term goal of limiting global temperature rise to well below $\degreeC{2}$ above pre-industrial levels \cite{deconto2021paris}. 
At the same time, it is important to maintain economic development, e.g., achieving growth and reducing inequality.
For example, international trade treaties, foreign investment, and technology transfer may enable developing countries to meet net-zero commitments, while contributing to global economic growth.
Such cooperation may be implemented through climate clubs~\cite{nordhaus2015climate}, which may overcome barriers to taking action on climate change.

From a modeling point of view, achieving and maintaining global cooperation poses a complex game-theoretic problem involving cooperation, communication, and competition.
This game can be modeled with $\numRegions{}$ strategic agents, each agent representing a region or nation.
Each agent is (boundedly) rational and implements policies to achieve its own socio-economic and climate objectives, which may be at odds with the objectives of the other agents. 
Agents interact through trade, diplomacy, or foreign aid and investments.
Cooperation may happen through mutual negotiation and agreements.

However, a crucial issue is that there is no central entity to enforce cooperation or adherence to signed agreements in the real world. 
For these reasons, it is critical to design multilateral negotiations and agreements that best foster \emph{continuing} cooperation towards mitigating climate change while allowing all parties to meet their individual policy objectives.

Such game-theoretic problems pose unsolved technical challenges. 
For instance, a key analysis in the 2022 IPCC report~\cite{portner2022climate} predicts climate change under five different so-called Shared Socio-Economic Pathways (SSP). 
Each SSP uses a manually defined set of climate policies for each global region. 
However, a key limitation is that it is unclear whether these policies would be executed by rational actors, and thus how likely these scenarios are to materialize or how robust these scenarios are to agents who may change their behavior over time.

\paragraph{Contributions.}
In this paper, we propose a conceptual and practical framework to overcome these limitations and provide a strategic analysis of rational agent behaviors and their impact on climate change.
We also introduce \emph{AI for Global Climate Cooperation}, a community initiative to foster interdisciplinary research and real-world impact in this area.
Specifically, we present the following:
\begin{itemize}
\item we argue that machine learning (ML) offers an attractive framework to analyze the strategic aspects of real world climate negotiations and agreements;
\item we argue this research requires combining both existing and new tools: calibrated agent-based climate-economic simulations, multi-agent machine learning, deep reinforcement learning (RL), game theory, and integrating data;
\item we urge the community to pursue interdisciplinary research, e.g., between economics, climate science, machine learning and other fields to understand the impact of strategic behavior on climate change;
\item we introduce \SimulationName{}, a calibrated climate-economic simulation that can model negotiations, agreements, and strategic behaviors between multiple regions;
\item we describe how to use multi-agent reinforcement learning (MARL) to train rational agents in \SimulationName{};
\item we introduce a working group collaboration to build momentum and to have actual policy impact through peer-reviewed publications authored by groups and policy briefs and communication with policymakers; and
\item as a first step, we are organizing a competition to evaluate solutions, both through quantitative results and scientific assessment by a diverse, interdisciplinary expert jury. 
The objective of this competition is to design contracts and protocols that foster global climate cooperation, using a mix of machine learning tools, economic intuition, and domain knowledge. 
Participants will evaluate their solutions using \SimulationName{} and a web platform run by the group.
 \end{itemize}
\textbf{For up-to-date information, see \url{www.ai4climatecoop.org}.}

\section{Related Work}
%

\paragraph{Integrated assessment models.}
Climate-economic simulations, such as \SimulationName{}, are known as Integrated Assessment Models (IAM).
IAMs are commonly used in climate change policy analysis to simulate future scenarios, based on a joint climate-economic dynamics model that quantifies the effect of economic activity and $\text{CO}_{2}$ emissions on global temperatures and long-term economic development.
A pioneering example is the Dynamic Integrated model of Climate and Economy (DICE)~\cite{nordhaus2007review}, which models the links among climate and economic factors, e.g., population growth, technological change, $\text{CO}_{2}$ emissions and concentrations, global temperatures, and economic damages.
Notably, DICE models a single global region. 
The Regional Integrated Climate-Economy (RICE) model generalizes this to multiple heterogeneous regions, modeling heterogeneous population growth, climate change mitigation investments, and other factors. 
The RICE model was further extended to include trade and tariffs \cite{lessmann2009effects}.


Key limitations of IAMs include a lack of modeling decision-making under uncertainty, distributional analysis, technological change, and realistic economic damage functions~\cite{farmer2015third}. 
Moreover, cooperation, communication, and competition between regions can significantly influence regional policies and their (long-term) effects, but are under-explored in IAMs. 
Our proposed framework, based on \SimulationName{}, machine learning, and other domains, addresses the latter shortcoming and is inspired by agent-based modeling~\cite{bonabeau2002agent} as a bottom-up modeling framework and advances in MARL to find well-performing policies~\cite{doi:10.1126/sciadv.abk2607} and to train human-level strategic agents~\cite{silver_mastering_2016,vinyals_grandmaster_2019}.


\paragraph{Multi-agent aspects of climate change.}
Previous work has studied the connection between political economy, negotiations, and climate change. 
Empirical work has found that previous climate summits have had inconsistent or too little impact~\cite{chan2022assessing, bakaki2022impact}.
The impact of social dynamics has been studied in a stylized climate-social model, finding that public perception and institutional responsiveness are important to explain variations in emissions~\cite{moore2022determinants}. 
Coalition-forming under climate negotiations and agreements have also been studied from a game-theory perspective ~\cite{zenker2019international}.
IAMs have also been used to study the impact of political bargaining on the economic burden required to meet climate targets~\cite{rochedo2018threat}.
However, to the best of our knowledge, no work has analyzed the game-theoretic aspects of climate cooperation using machine learning and calibrated IAMs.

\paragraph{Strategic behavior and climate change.}

Game theory has long studied the collective behavior of self-interested agents, e.g., the tragedy of the commons~\cite{hardin1968tragedy}, negotiation and agreements of agents with conflicting and common goals~\cite{schelling1980strategy}.
Certain works have analyzed international negotiations on climate collaboration and agreements regarding economic activity and climate efforts, e.g., imposing tariffs on countries that do not mitigate sufficiently.
Climate negotiations have been studied using mathematical games, e.g., coordination games or prisoner's dilemmas~\cite{decanio2013game}. 
However, the reliability of such simplified models for real-world policy has been called into question.
In particular, these games lack 
(i) a multilateral, rather than bilateral, setting, 
(ii) strategic behavior from agents with multiple, possibly conflicting, goals, 
(iii) evolving climate dynamics and changing agent behavior that lead to non-equilibrum outcomes, and 
(iv) heterogeneity among agents~\cite{madani2013modeling}.

Subsequent work has gone beyond equilibrium analysis by modeling climate negotiations as a bargaining game in which agents learn, albeit in a highly simplified manner~\cite{smead2014bargaining}.
Climate scenarios could be studied through the emergence of climate mitigation from these games with learning, in which regions can cooperate or compete~\cite{greeven2016emergence}.
Furthermore, other work has studied the difficulty of long term climate collaboration~\cite{carney2015breaking}, as well as potential mechanisms for overcoming associated issues~\cite{nordhaus2015climate}.


\paragraph{Multi-agent reinforcement learning.}
MARL has emerged in recent years as an attractive framework that studies how to train rational agents that may communicate, cooperate, or compete. 
This is a rich area of research that intersects machine learning with game theory, economics, and other domains~\cite{shoham2008multiagent}.
Games can be classified as cooperative, competitive, or a mixture of both. 
For instance, in fully cooperative games, agents learn to work together, e.g., to lower the carbon power consumption of heating, ventilation and air conditioning (HVAC) systems ~\cite{mai2021multiagent, https://doi.org/10.48550/arxiv.2110.13450, https://doi.org/10.48550/arxiv.2006.14156}, or in the game of Hanabi ~\cite{https://doi.org/10.48550/arxiv.2103.01955}. 
On the other hand, in a competitive game agents may need to find strategies to defeat opponents, e.g., in Diplomacy ~\cite{NEURIPS2019_84b20b1f} and Go~\cite{Schrittwieser_2020}. 
However, many games are neither purely competitive nor purely cooperative. 
Hence, multi-agent games can be simple to define, yet hard to find optimal agent strategies using learning-based methods, e.g., in the iterative prisoner's dilemma and coin games ~\cite{https://doi.org/10.48550/arxiv.1707.01068, lola}.

Recent work has explored the link between MARL and negotiation~\cite{cao2018emergent}, as well as cooperation in social dilemmas and collaboration on climate change~\cite{jaques2019social,chelarescu2021deception,le2022towards}.
As such, MARL is an attractive framework to analyze climate outcomes taking strategic behavior into account. 
However, previous work has largely considered highly stylized environments and has not yet been applied to rich calibrated climate-economic simulations; our work fills this gap.

\section{The \SimulationName{} Integrated Assessment Model}
\label{sec:ABM}

\begin{figure*}[ht!]
    \centering
    \includegraphics[width=\linewidth]{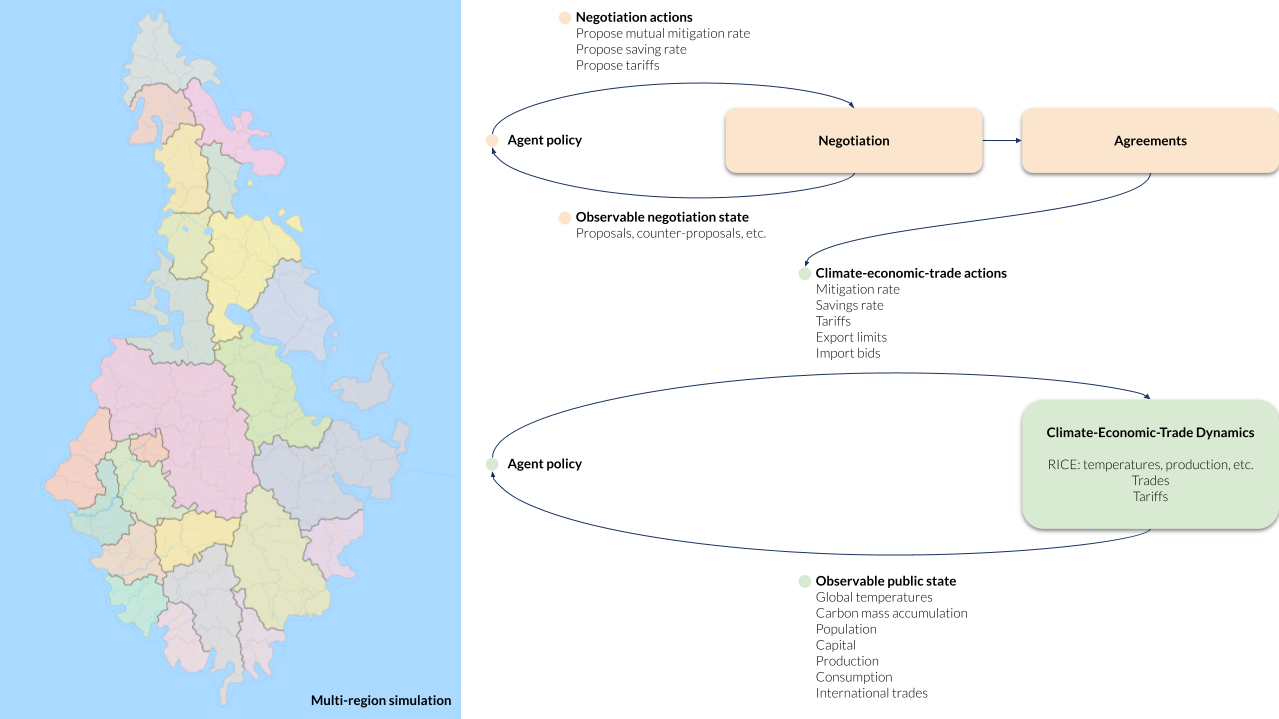}
    \caption{\textbf{Schematic overview of observables, actions, and components in \SimulationName{}.} 
    Each region (agent) uses a policy model to negotiate and make climate, economic, and trade decisions. 
    For clarity, we show the flow of information and actions for a single agent only.
    The details of the negotiation protocol are up to competitors to implement.
    At a high-level, at each timestep, each policy first receives an observation of the negotiation and world state and decides how to negotiate.
    Negotiation may proceed for several iterations (at the same timestep).
    The outcome of all negotiations are agreements (or lack thereof), which may be between two or more agents. 
    In particular, an agreement may influence the remaining actions that an agent can take in the climate and economic domains.
    For the same timestep, each agent then makes decisions with respect to the climate, economy, and trade. 
    }
    \label{fig:simulation-schema}
\end{figure*}

We introduce \SimulationName{}, an IAM that further augments RICE with a framework for negotiation protocols, and also includes international trade and tariffs, following \citet{lessmann2009effects}).
As such, \SimulationName{} shares climate, economic and social characteristics with the real world.

In \SimulationName{}, there are $\numRegions$ regions, each modeled as an independent decision-making agent.
Regions interact with each other and the environment through their actions: setting a savings rate, mitigation rate, trades and tariffs, and negotiation actions, for each time step (e.g., every 5 years).

\SimulationName{} has two main components: \emph{negotiation} and \emph{climate-economic activity}, see Figure~\ref{fig:simulation-schema}.
The activity component simulates the physical actions of the agents and the resulting evolution of the environment.
The negotiation component simulates communication between regions, allowing them to influence each other's behaviors and form agreements.
Agreements may in turn adjust the available actions for each region during the activity stage.

Each simulation episode consists of $\numEpisodeSteps$ steps, each step representing $\numStepPeriod$ years (e.g, $\numStepPeriod=5$).
Thus, the simulation spans a period of $\numEpisodeSteps \times \numStepPeriod$ simulation-years.
At every step, the simulation goes through the negotiation stages, and agreements are formed within the regions.
The simulation then enters the activity stage where each region takes actions that are affected by the agreements formed during the negotiation stages.

\emph{
For the public version of the simulation, we have fitted the structural parameters of the simulation to real data.
However, the regions and their characteristics in the public version are not an identical one-to-one representation of the real world. 
Moreover, the regions are fictitious and do not represent real-world nations.
}

\subsection{Base Dynamics}

We now describe the climate-economic and trade dynamics of the simulation that are based on existing work.

\paragraph{Climate-economic dynamics overview.}

The \emph{state of the world} is characterized by global variables such as the concentration of $\text{CO}_{2}$ levels in the Earth's atmosphere, and the average global temperature, as well as region-specific variables such as population, capital, technology level, carbon intensity of economic activity, and balance of trade (see Table~\ref{tab:world_state_var} in Appendix J).

The climate dynamics model how $\text{CO}_{2}$ levels in the atmosphere impacts global temperatures. 
The economic dynamics model how technology levels, capital, population, and gross domestic production evolve. 
Notably, the climate dynamics impact the economic dynamics through a \emph{damage function} which describes how higher temperatures lead to losses in capital.

These dynamics depend on savings and mitigation rates set by each agent, e.g., agents may choose to invest more in climate change mitigation, but this may lower economic productivity in the short-term.%
\footnote{
In economic terms, variables such as capital, balance of trade, carbon mass, and global temperature depend on the agents' actions and are called \emph{endogenous} variables.
On the other hand, variables such as population, technology level, carbon intensity of economic activity are called \emph{exogenous}, i.e., their values do not depend on the agent actions.
Note that the values of endogenous variables can vary across steps in a predetermined manner.
}
As global $\text{CO}_{2}$ levels and temperatures affect all agents, these dynamics mean the decisions of each agent affect the climate-economic outcomes for other agents too.

The activity component encapsulates these dynamics.
At every step, it does the following: 
The gross output production for each region is computed based on the state of the region, in particular, its capital investment, labor (or population), and technology factor. 
The net economic output is the gross output production reduced by climate damages from rising global temperature, and the cost of efforts towards mitigation by this region.
The region consumes domestic goods equal to the quantity of the net economic output that is left after capital investment and export.
It also consumes foreign goods from imports.
The consumption utility for each region from consuming domestic and foreign goods is computed using the Armington elasticity assumption that has become standard in international computable general equilibrium models \cite{armington1969theory}.
This gives the \emph{reward} corresponding to each region in every step. 
Please refer to Appendix \ref{sec:climateecon_appendix} and Table~\ref{tab:agent_action_var} in Appendix J for more details. 

\paragraph{International trade and tariffs.}
\SimulationName{} features international trade to exchange and transfer goods between agents, following~\cite{lessmann2009effects}.
Here, agents are modelled to seek diversity in their consumption according to the Armington assumption~\cite{armington1969theory}, so they want to consume goods produced by other agents and are willing to export some of their own goods in exchange.
Each agent specifies how many goods they would like to import and sees its orders (partially) filled, depending on the other agents' willingness to export.
%

In addition, agents can choose to impose import tariffs to restrict trading.
Import tariffs restrict the consumption of the imported goods that they are applied to, which implicitly increases their prices.
These price increases make imported goods without import tariffs more attractive. 
Therefore, trade and tariffs force agents to engage with other regions and be strategic, and hence may incentivize negotiations and signing agreements.
Please refer to Appendix \ref{sec:trade_appendix} for more details.

\section{The Negotiation Component}
\label{sec:model_negotiation_framework}

\emph{A key objective of our working group collaboration and competition is to test different negotiation protocols that affect the actions during the activity stage and lead to socially better outcomes.}
As such, the negotiation component can include different stages of communication and the rules to form agreements between the regions.
Note that \SimulationName{} can work with an empty negotiation component, and indeed these provide important baselines for performance evaluation.

\emph{%
For a detailed exposition and formal details on negotiation protocols and agreements, see Section \ref{sec:negotiation-formal-details} in the Appendix.
}


\paragraph{Binding agreements via action masks.} 
In the base implementation of \SimulationName{}, we allow the negotiated agreements to control the range of actions allowed for the regions.
%
%
The effect of an agreement can then be encoded in the form of \emph{action masks} that controls the allowed action space during the activity stage.
For example, an agreement could state that a given region should implement a minimum of $20\%$ mitigation rate.
Then the action mask only allows setting mitigation rates above that level for this region.

Hence, it is important to ensure that the negotiation protocol provides each region with a way to reject any agreement that restricts its actions.
For example, it shouldn't happen that a certain region is forced to implement a mitigation rate that it did not agree to~\cite{mitchell2003international}.
Even if the agents are provided a choice to accept or reject an agreement, action masks could be implemented by the simulation designer to artificially restrict actions to make an agreed-upon agreement fully-binding.
However, this is less realistic.

\paragraph{Non-binding agreements.}
Alternatively, an implementation can include agent observations that allow regions to check if the agreements were followed or not and learn to punish the regions that did not comply with them.
Alternatively, a reputation system could be built based on the observations corresponding to agents complying with their agreements, with agents using these reputation scores to engage with each other for future agreements.
Such implementations would be more realistic as compared to an implementation based on action masks because they do not assume any agreement enforcing entity.

\paragraph{Properties of negotiation protocols.} 
A wide range of protocols can be implemented as part of the negotiation component. 
A key goal is to design a negotiation protocol so that agents may not only maximize their own utility but also care about the collective goal: mitigating global temperature rise.
From a theoretical point of view, attractive properties of negotiation protocols could be:
\begin{itemize}
    \item \emph{incentive-compatible}: every agent can achieve the best outcome for themselves by acting according to their true preferences~\cite{pavan2014dynamic},
    \item \emph{self-enforcing}: no external body or agent is required to enforce other agents to participate in negotiations and adhere to agreements~\cite{telser1980theory},
    \item \emph{strategy-proof}: without information about other agents' actions, agents do at least as well by being truthful~\cite{li2017obviously}.
\end{itemize}
We now discuss several example negotiation protocols.

\subsection{No Negotiations}
\label{sec:sample_simulation_no_negotiation}

Let us begin with the most simple negotiation protocol, the one with no negotiation.
In this case, the simulation runs through the activity component at each step.
Each region chooses its actions from the entire feasible range of actions without any restriction, since no agreements are formed.
Each region aims to optimize for its individual rewards and thus this is the classic case of tragedy of the commons.


\subsection{Unilateral Contracts}
The negotiation component could include various constraints that do not require negotiation.
For example, if two regions $a$ and $b$ enact mitigation rates $\mu_a$ and $\mu_b$ such that $\mu_a > \mu_b$, then an agreement could entail that region $a$ imposes a tariff $\tau_a = \alpha(\mu_a - \mu_b)$ on region b where $\alpha$ is a ``mitigation correction'' coefficient.
This is a simple mechanism through which a region $a$ could incentivize regions $b$ to mitigate more.
%

\subsection{Bilateral Negotiations on Mitigation Rates}
\label{sec:bilateral negotiations}

The negotiation component can be composed of multiple stages of observations and actions that ultimately lead to agreements within the regions.
Consider the following protocol:
For each ordered pair of regions $(\idxi,\idxj)$, region $\idxi$ makes a proposal $(\hat \xmitigation_\idxi, \hat \xmitigation_\idxj)$ and region $\idxj$ decides whether to accept the proposal or not.
The above proposal means that, if accepted by region $\idxj$, 
then an agreement is formed between region $\idxi$ and region $\idxj$ that says, region $\idxi$
will choose a mitigation rate $\xmitigation_\idxi$ 
at least as large as $\hat \xmitigation_\idxi$ and region $\idxj$ will choose a mitigation rate $\xmitigation_\idxj$ 
at least as large as $\hat \xmitigation_\idxj$ 
as their inputs to the activity component.

Thus, during negotiation, there will be $\numRegions(\numRegions-1)$ total proposals and decisions regarding their acceptance.
After negotiation, each region $\idxi$ will select its mitigation rate $\xmitigation_\idxi$ that is greater than or equal to all of the mitigation rates $\hat \xmitigation_\idxi$ in the accepted proposals that region $\idxi$ was a part of.

This negotiation protocol can be implemented in two stages:
\begin{itemize}
    \item \emph{Proposal stage}: At this stage, each region $\idxi$ makes a proposal to every other region $\idxj$.\\
    Observations: state observations for agent $\idxi$.\\
    Actions: Proposals $(\hat \xmitigation_\idxi, \hat \xmitigation_\idxj)$ for every other region \idxj.
    \item \emph{Evaluation stage}: At this stage, each region observes the proposals made to it in the preceding stage and takes an action of accepting or rejecting each of the proposals.\\
    Observations: state observations for agent $\idxi$, incoming and outgoing proposals for agent $\idxi$.\\
    Actions: Accept or reject each received proposal.
\end{itemize}

At the end of the evaluation stage, the minimum required mitigation rate is computed for each region and the corresponding action mask is set.
This ensures that the regions choose actions for the activity component that comply with the agreements formed in the negotiation component.
Instead of using action masks, the implementation can allow agents to take any action, but include these actions as observations for the other agents so that they can learn to punish any agreement violations.
For example, regions can increase the tariff rates on regions that violate agreements.

\begin{figure*}[ht!]
    \centering
    \includegraphics[width=\linewidth]{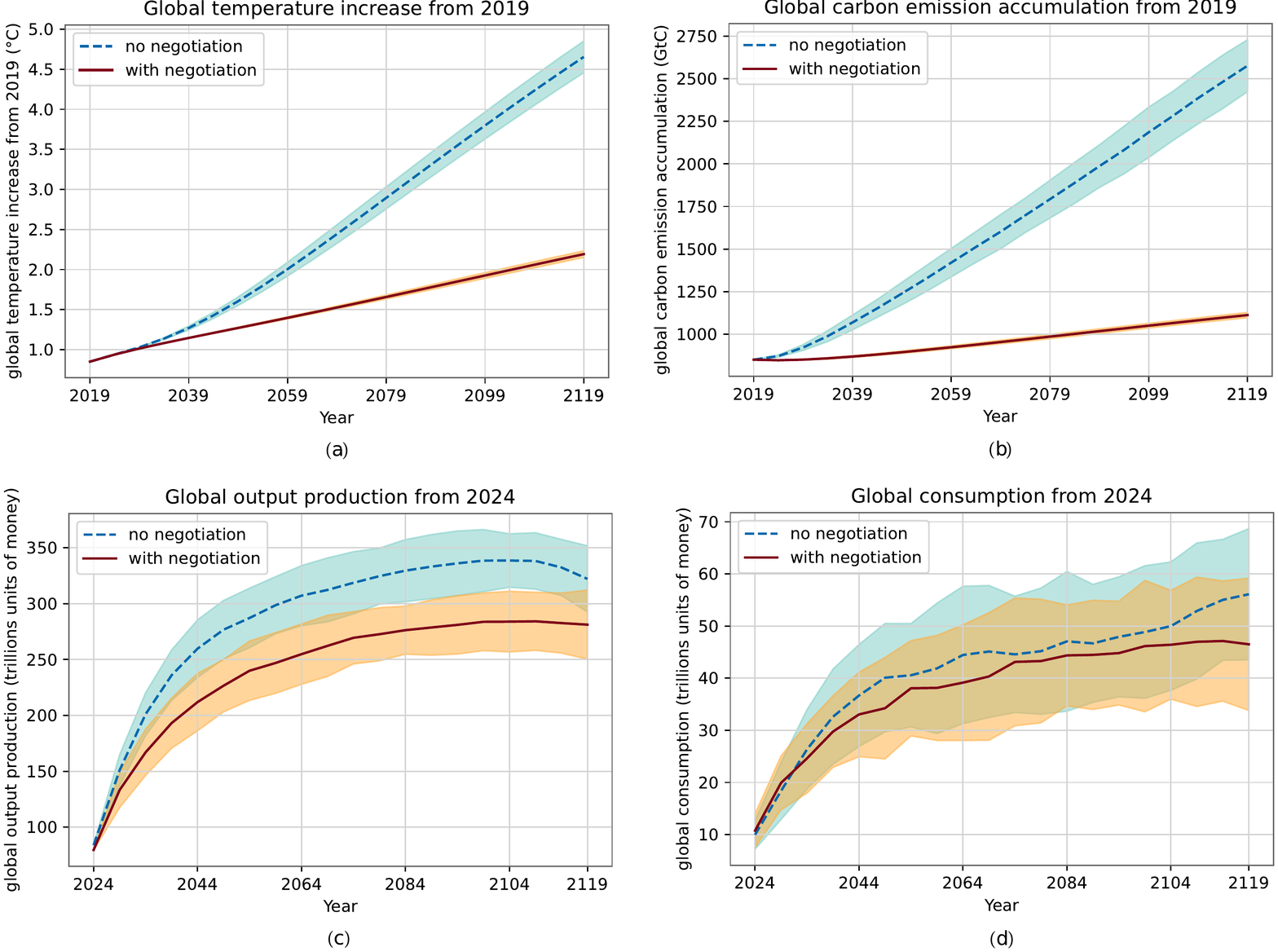}
    \caption{\textbf{Comparing outcomes with RL agents trained with and without a simple bilateral negotiation protocol.} 
    Here we show the simulated outcomes under the simple bilateral protocol from Section \ref{sec:bilateral negotiations}, and agents that were independently trained using PPO.
    (a) The predicted increase in global temperature, with predictions starting in 2024 and initialization in 2019.
    (b) The predicted global carbon emission accumulation, with predictions starting in 2024 and initialization in 2019.
    (c) The predicted global output production (aggregated over regions) starting from 2024.
    (d) The predicted global consumption (aggregated over regions) starting from 2024. 
    The solid blue and red lines represent the mean across 100 repetitions with random seeds without negotiation and with negotiation respectively;
    the shading represents 1.96 times the standard error. 
    In each simulation episode, agents action are decided on and executed every 5 years.
    }
    \label{fig:simulation-results-under-negotiation}
\end{figure*}

\paragraph{Visualizing outcomes.}

Figure \ref{fig:simulation-results-under-negotiation} visualizes the effect of a baseline implementation of the bilateral negotiation protocol, with the agents trained independently using PPO~\cite{schulman2017proximal}, an RL algorithm (see Section \ref{sec:AgentModelsUsingML} for details). 
You can tell that the negotiations and agreements have a beneficial effect: global temperatures rise less than without negotiation. 
However, note that this baseline protocol may not achieve the best trade-off in climate-economic outcomes possible.
Also note that this protocol are not necessarily robust against sim-to-real gaps, e.g., if the agent reward function is mis-specified or the dynamics are not calibrated well. 
A key objective of this initiative is to investigate such protocols in more detail and elucidate their properties.


\subsection{Multilateral Negotiations}


Multilateral negotiations imply communication amongst several regions simultaneously. 
Although more technically challenging to implement, such negotiation protocols offer more possibilities and added realism.
A key goal of this competition is to explore and implement effective negotiation protocols. 
We now discuss a recent proposal as an example.

\paragraph{Climate clubs.}
Climate clubs \cite{nordhaus2015climate} are one example of a multi-region agreement. 
Each region that is a part of a club agrees to 1) enact a minimum mitigation rate and 2) enforce a minimum import tariff on regions that are not part of the club.
A high minimum import tariff acts as an incentive to join the climate club by applying a mitigation rate at least as high as the club's minimum.

The negotiation protocol would specify how clubs are formed, e.g., by randomly selecting one region to make a proposal $(\mu_{min}, \tau_{min})$ to which all other regions can agree (thereby joining the climate club) or refuse.
This could be augmented in many ways, e.g., by repeating it with the standalone regions until all regions form a climate club, by implementing an iterative proposal-evaluation framework, or by specifying a more sophisticated mechanism for climate club proposals than random selection.

\paragraph{Asymmetric negotiation protocols.}

Certain negotiation protocols, e.g., climate clubs, rely on the uniform treatment of trading partners to simplify their theoretical analysis~\cite{nordhaus2015climate}. 
However, such limits do not apply to agent-based simulations~\cite{farmer2015third}, and therefore the negotiation actions of regions in the simulator are not required to be applied uniformly across trading partners.

In fact, effective international cooperation is based on \emph{climate justice}~\cite{klinsky2017equity}, which often requires that countries be treated differently based on their characteristics, e.g., historical cumulative emissions or exposure to potential climate damages~\cite{declaration1992rio}.
For example, article 4.4 of the Paris Agreement states that developed regions, which have historically contributed emissions in a disproportionate manner relative to developing countries, should continue taking the lead when it comes to mitigation efforts~\cite{schleussner2016science}.
However, there is no agreed-on way to label regions as ``developed'' and ``developing'' in simulations such as \SimulationName{}, nor is there unanimous agreement on the validity of that particular denomination.
Rather, we invite the community to consider how the characteristics of the fictitious regions in \SimulationName{} relate to the climate debate in the real world.

Moreover, the ``fairness'' of climate change legislation can be established through many different methodologies which treat regions differently~\cite{dooley2021ethical}. 
We encourage the community to explore important opportunities for the innovation and study of negotiation protocols that implement forms of climate justice.

\emph{Although building any simulation requires making design choices, we emphasize that we do not seek to make normative statements about ``fairness'' and ``justice'' in this work.}
\section{Modeling Agent Behavior using ML}
\label{sec:AgentModels}
\label{sec:AgentModelsUsingML}

The negotiation protocols and the climate-economic dynamics of RICE-N 
define a game-theoretic setup between the different regions.
We model the behavior of an agent $\idxi$ using its policy $\policy_\idxi\brck{\ac_\idstep|\ob_\idstep}$ 
that maps the agent's observations to a probability distribution over its actions.

\paragraph{Why use ML?}
Existing IAMs often use a predetermined policy for each agent that is fixed exogenously.
Such an approach would require handcrafted agent policies for each different negotiation protocol
and the reliability of the outcomes from the simulation will depend on the modeled policies.

In contrast, we assume each region to be a strategic agent that is interacting with the environment and other strategic agents in it.
Here, we do not manually set the behavioral policy $\policy_\idxi$, rather we use machine learning techniques to find the policy that optimizes the agent's objective, and hence derive the agent policies endogenously.

Specifically, the agents are assumed to be rational such that each agent $\idxi$ 
optimizes its policy $\policy_\idxi\brck{\ac_\idstep|\ob_\idstep}$ 
to maximize its long-term aggregate $\df$-discounted utility:
%
%
\eq{\label{eq:agent-opt}
\max_{\policy_\idxi} \; &\E_{\policy_1, \ldots, \policy_\numRegions}\brcksq{\sum_{\idstep=0}^\numEpisodeSteps \df^\idstep\rew_{\idxi,\idstep}}.
} 
Here, $\rew_{\idxi, \idstep}$ is the utility of the region $\idxi$ at step $\idstep$ determined by its aggregate consumption $\xconsumption_{\idxi, \idstep}$ as follows:
\eq{ \label{eq:utility}
\rew_{\idxi, \idstep} = \xutility_{\idxi, \idstep} = \frac{1}{1-\alpha} \xpopulation_{\idxi, \idstep} \brck{\frac{\xconsumption_{\idxi, \idstep}}{\xpopulation_{\idxi, \idstep}}}^{1-\alpha},
}
where $\xpopulation_{i,t}$ is the population of the region $\idxi$ at step $\idstep$.
The discount factor $\df$ models the long term value of rewards for the agents and as such they could differ across different agents.
The discount factor for each region is often updated with the changing administration and in the absence of any consensus we fix $\df$ to be homogeneous across agents as assumed in \cite{nordhaus2015climate}.
The aggregate consumption $\xconsumption_{\idxi, \idstep}$ in the above equation is obtained by combining the domestic and foreign goods consumption using the Armington model \cite{armington1969theory}:
\eq{
    \xconsumption_{\idxi, \idstep} &= \brck{  \psi^{\textrm{dom}}(\xconsumption_{\idxi, \idxi, \idstep})^\lambda +  \sum_{j\neq i} \psi^{\textrm{for}} (\xconsumption_{\idxi, \idxj, \idstep})^\lambda  }^{\fr{1}{\lambda}}, \\
    \xconsumption_{\idxi,\idxj,\idstep} &= \xgoodstraded_{\idxi,\idxj,\idstep} (1 - \xtariff_{\idxi,\idxj,\idstep}) \quad \forall j\neq i,
} 
where $\xconsumption_{\idxi,\idxj,\idstep}$ are the foreign goods consumed after imposing tariffs $\xtariff_{\idxi, \idxj, \idstep}$ on the imported goods $\xgoodstraded_{\idxi, \idxj, \idstep}$ by region $\idxi$ from region $\idxj$ at step $\idstep$. 
Furthermore, $\psi^{\textrm{dom}}, \psi^{\textrm{for}}$ are shared parameters and $\lambda$ is the Armington elasticity parameter.



\paragraph{Multi-agent reinforcement learning.}
%
Simulating a rational agent requires computing the optimal policy for each agent.
\footnote{Although we assume the agents to be rational, other behavioral models could 
also be implemented in \SimulationName{}.}
Finding the optimal rational policy for each agent in response to complex environment dynamics and other agent policies naturally leads to MARL; \citet{busoniu2008comprehensive} give a comprehensive overview of the topic.
In short, MARL extends single-agent RL to find 
an optimal policy for each agent interacting in a dynamic environment to solve Equation \ref{eq:agent-opt}.

The RL framework models how an agent's actions affect the state of the environment 
and its rewards (utilities).
Thus an RL agent has to learn to anticipate the long-term effects of its actions.
This is especially true and challenging in multi-agent environments, e.g., in \SimulationName{}, where agent actions affect key climate-economic metrics, e.g., global temperatures, capital investments, etc.
In addition, MARL solution algorithms have to deal with additional challenges since each agent has to respond to the policies of other agents. 
This makes the task of finding the optimal policy a moving target (until a form of equilibrium is reached in the agent policies).
Developing effective MARL is an ongoing research topic, and we encourage the community to explore MARL as a solution framework to analyze the short and long-term effects of (sub-)optimal behavior on the climate and economy.

\paragraph{Implementing RL agents.}
\label{sec:RLagents}
Our public code~\cite{Zhang_RICE-N_2022} includes both CPU and GPU implementations of the full RL pipeline using two popular RL algorithms: A2C and PPO \cite{schulman2017proximal}. 
\SimulationName{} can also be used with other RL implementations.

Our base implementation models each RL agent using a neural network policy that shares weights across agents, but uses region-specific inputs.
The architecture of the network can be adjusted, e.g., the number of layers and the dimension of each layer.
Agent policies use separate heads for each action.
To distinguish between agents, the policy model's input contains agent-specific features, e.g., their population, capital, technology factor, damage function, and a one-hot representation of the region's index, as well as the public state of the world (e.g., climate conditions).

In addition, each agent gets information about negotiations, e.g., the latest proposals made to and by this region, or the minimum mitigation rate agreed upon by this region.
How negotiations evolve depends on the specifics of the protocol and the different actions executed by the agent, e.g., proposals for other regions, decisions on proposals made by other regions, and setting mitigation and savings rates that may or may not be in line with what was agreed on.
Depending on the negotiation protocol, not all observations and actions are relevant to each agent.
Hence, it is recommended to use efficient implementations that only feed the information to agents that are relevant at a given time.
\section{Evaluating Negotiations and Agreements}
\label{sec:evaluation_criteria}

\paragraph{Evaluating solutions.}

A complete solution consists of a variation of \SimulationName{} with a negotiation protocol, agreements, and agents (trained using RL or whose behavior is learned or defined through other means). 
We can then measure the performance of a solution via global economic and climate metrics. 
In particular, because there may be intrinsic trade-offs between these objectives, we are interested in the set of best trade-offs, i.e., the \emph{Pareto frontier} that are achievable under a solution framework. 
Hence, we focus on estimating the area-under-curve of the Pareto frontier that is achievable for a given solution.

\emph{
We define performance in this way, because the framework itself should not encode a preference for a certain set of climate and economic outcomes. 
Which specific set of outcomes is desired represents a social choice; in this work, we do not intend to make any normative statements about which specific outcome is preferred.
}

\paragraph{Objectives.}
The climate index $\climateindex$ measures the temperature change over the course of 100 years, \emph{compared to two extremal policies that do not mitigate at all or use a 100\% mitigation rate}.
The economic index $\economicindex$ measures the increase in global productivity\footnote{Exploring additional, more granular measures of economic well-being within the simulator is an interesting area of exploration. } (i.e., total GDP), similarly compared to two extremal policies.
Each index is computed as a relative gain compared to extremal agent policies to get normalized, dimensionless metrics along both climate and economic dimensions. 
This allows for fair comparison and computation of an area-under-curve metric in the $\brck{\climateindex, \economicindex}$-space.
A basic intuition is that these two extremal policies represent two extremal points (very high $\climateindex$, very low $\economicindex$, and vice versa).
However, these extremal policies may not be Pareto-optimal, may not be optimal under every negotiation protocol, and may not be individual best-response strategies.

\paragraph{Empirical Pareto frontier.}

Each solution framework may generate a set of different outcomes $\brck{\climateindex, \economicindex}$, e.g., if the negotiation protocol has a free parameter that may change how agents behave (optimally).
In addition, each specific outcome needs to account for randomness by averaging over Monte-Carlo roll-outs performed under different random seeds.%
\footnote{
For example, the base implementation of \SimulationName{} has deterministic climate-economic dynamics, but the agents' behavior may be stochastic. 
In our competition, we use 10 simulation roll-outs under 3 random seeds, and determine the mean and standard deviation of the outcomes for evaluation purposes.
}

Each solution thus yields a set of points in the $\brck{\climateindex, \economicindex}$-space, from which we can infer an empirical Pareto frontier: the set of points for which neither index can be increased without decreasing the other index. 
Specifically, a point $a$ \emph{Pareto-dominates} another point $b$ if at least one index is strictly higher and the other index is at least as high, i.e., 
1) $\climateindex_a > \climateindex_b$ and $\economicindex_a \geq \economicindex_b$, or 
2) $\climateindex_a \geq \climateindex_b$ and $\economicindex_a > \economicindex_b$.

The empirical area-under-curve is defined using a \emph{hypervolume indicator}: the area of a set of non-dominated policies in the objective space with respect to a predetermined reference point~\cite{van2014multi}.
In our case, the reference point is set by the extremal policies.

Besides giving a numerical value to compare different solutions in a multi-objective setting, this score definition has several attractive properties in a competition setting:
1) The hypervolume indicator encourages competitors to lie as close as possible to the true Pareto frontier,
2) it is strictly monotonic with respect to Pareto dominance such that a set of points that Pareto-dominates dominates another set must have a bigger hypervolume, and
3) it incentivizes diversification (along the Pareto frontier), i.e., competitors are incentivized to submit solutions that make different climate-economic trade-offs to increase the hypervolume of their set.

See Appendix \ref{sec:submission-evaluation-details} and Figure \ref{fig:hypervolume} for more details.

%
%

\section{Engineering aspects and customization}

The \SimulationName{} simulation builds on the RICE model \cite{nordhaus2015climate}, a multi-region climate-economic simulation, by adding negotiation protocols, international trade, and support for strategic agents. 
\SimulationName{} has $\numRegions$ agents, each representing a region in the world or a group of (fictitious) countries that are assumed to make decisions as a single entity.
To maintain realism, we calibrate the structural parameters and establish a reasonable range of agent-specific parameters using data from the World Bank API \cite{wbapi}. 
Algorithm \ref{alg:cap} in Appendix \ref{sec:rice-dynamics-details} shows the full flow of the activity component.

\paragraph{Code modularity.}
\SimulationName{} is designed to be easily extendable and modular. 
In particular, \SimulationName{} provides a framework to implement negotiation protocols as well as learning algorithms (e.g., RL agents) to enable designing negotiation protocols and study their effects on global cooperation.

The simulation code is structured so that custom negotiation protocols can be implemented as part of the negotiation component.
An implementation may add agent observations and actions corresponding to the proposed negotiation protocol, 
and rules to generate the corresponding agreements.
Notice that the agents can have access to these additional observations even during the activity stage.
These observations are useful in letting agents adopt policies in response to the negotiations and agreements.
For example, they can learn to punish agents that do not comply with the agreements. 
Besides, by using action masking, the simulation can ensure that the agreements are followed (we can also have mix where some agreements are enforced using action masking and some others are non-enforceable).
Thus, the simulation code provides complete freedom in implementing any negotiation protocol of choice provided the corresponding agreements are appropriately connected to the activity stage action through additional observations or action masks.
%


\paragraph{Consistency checks.}
On the other hand, for the purpose of the competition, we fix the actions in the activity stage and the corresponding climate and economic dynamics.
Thus, certain parts of the simulation should not be modified by competitors, e.g., the core climate and economic parameters and equation, such as the ones which affect carbon emissions and productivity.
Changing these components may change the problem and prevent fair comparisons across submitted solutions.
As such, our evaluation protocol and implementation of \SimulationName{} include consistency checks that test submitted solutions; each submission requiring submitting the full code of the (modified) simulation and agent models.
The activity stage is designed to follow 
the well established climate-economic dynamics in the RICE model 
augmented with trades and tariffs to include an essential influence in international negotiations.
This way the simulation follows a simple model but is rich enough to implement  interesting negotiation protocols.

Further, our implementation is set up to model each region as an RL agent that optimizes its discounted long-term rewards.
Although other agent policy models are possible, we require that the competitors clarify how their proposed agent policy models also aim to optimize for the same goal, namely, the personal discounted long term reward for each agent.
This ensures that the agents behave greedily and any improvements in global metrics are strictly a result of the implemented negotiation protocols.
See section \ref{sec:AgentModels} for details.

\section{Working Group and Competition}
To build momentum for this research direction, we invite the community to join our working group collaboration and competition.
The goal of this initiative is to help discover innovative, AI-based approaches to international climate negotiations that (i) go beyond traditional climate-economic trade-offs, (ii) are sufficiently sophisticated to have real world relevance, and (iii) are useful and relevant to policymakers.
Specifically, to motivate rapid development and comparison of different solutions, we are organizing a competition in which participants use RICE-N as a test bed to develop and evaluate negotiation protocols.

The working group consists of: 1) organizers of the competition, 2) competition participants, 3) an expert jury, and 4) advisors.
A concrete outcome and goal of this working group is to report the findings of the competition (vetted by an expert jury and internal review) in a peer-reviewed publication, co-authored by the competition participants and other working group members.
The findings would then be used to make policy and protocol recommendations to policymakers. 
This represents a first step toward the development of effective international climate agreements in the real world.

The competition has three tracks. 
The first track quantitatively evaluates the performance of a negotiation protocol on climate and economic performance.
The second track, in addition to the quantitative evaluation, involves a scientific assessment of feasibility by a diverse, interdisciplinary jury with expertise on machine learning, economics, climate science, computational social science, policy writing, international relations, law, and ethics.
Negotiation protocols championed by such a diverse jury are likely to be more viable for real-world application and suitable for policymaking.
The third track invites critical analysis and suggestions for improvements. 

\section{Discussion}
We presented the our working group collaboration, competition, the \SimulationName{} simulation, and a conceptual framework to include strategic negotiation into climate-economic modeling. 
To our knowledge, this is the first simulation and learning-based framework to study negotiation in climate-economic systems of this complexity.
Given the modular structure of our implementation, \SimulationName{} can be easily extended to include more climate, economic, or strategic features.
As such, we believe this is a compelling step to develop AI for climate change and invite the community to build on this work.

The RICE climate and economic dynamics are relatively simple, but are a minimal implementation of an IAM that supports studying the strategic aspects of decentralized decision-making, negotiation, and their impact on the climate. 
Both the climate and economic components could be made more sophisticated, as implemented by a large number of existing IAMs.
For example, one might add rich spatial climate dynamics, or model more parts of the real economy, e.g., international companies or markets with price discovery.
The game-theoretic features could also be enhanced, e.g., by allowing for forms of communication, more policy levers, more possible interactions (e.g., technology transfer), and others.
The behavior models for the agents could also be modified, for example, by considering heterogeneous discounting coefficients, alternative utility functions with different shapes, etc.
Besides, instead of considering a time span of 100 years for each agent's optimization, we can consider an election cycle, thus making the agent incentives more relevant from a practical point of view.
However, simulating more complex dynamics also incurs significant computational cost.
As such, we hope that future work will extend the \SimulationName{} model towards using AI for climate change mitigation and social good.
\section{Ethical Considerations}
\label{sec:ethics}

While the intention of this paper and the corresponding challenge is to stimulate innovative solutions to climate change, there are some unintended consequences that we would like to acknowledge and address here. These include the carbon footprint of running the simulation itself, the economic disparities that can exist with climate negotiations, and the potential extensibility of this simulation to the real world.

\paragraph{Carbon emissions from simulations.}
First, it is important to acknowledge that running climate change simulations in \SimulationName{} will inevitably release carbon emissions into our atmosphere. While the computational requirements of these simulations are much smaller than training large language models, they still exist. To mitigate this harm, we encourage participating teams to consider their energy use during experimentation, offsetting their carbon emissions if possible.
For reference emission estimates of our code and suggestions for measurement tools, please see Appendix \ref{sec:CO2-emissions}.


\paragraph{Economic and climate disparities.} 
Second, as the World Bank states ``Climate change is deeply intertwined with global patterns of inequality'' and yet ``the most vulnerable are often also disproportionately impacted by measures to address climate change'' \cite{wbsocialdimensions}. 
While it is important to determine ways that to mitigate climate change, it is equally important to ensure that vulnerable populations are not negatively impacted by climate change measures.

\paragraph{Technical aspects and limitations.}
Last but not least, it is important to note that the predicted climate and economic predictions made in \SimulationName{} may differ in a real-world setting due to externalities beyond the boundaries of the simulation. 
A fictional world is utilized in this competition to further illustrate the potential gap between simulation and reality, but the uncertainty of the results should be fully understood, especially before implementing any policies recommended by \SimulationName{}. 

\bibliography{references}  

\newpage
\appendix
\section*{Acknowledgements}
We thank Vincent Conitzer, Richard Socher, Nicholas Kumbleben, Christian Schroeder de Witt, Eyck Freymann,  and Catherine Labasi-Sammartino for valuable discussions and feedback.
In particular, we acknowledge Richard Socher for initial discussions about organizing an AI competition.
We thank Anna Bethke for the ethical review.
We thank Qianyi Cheng and Zhu Zhu for the competition logo design.
We thank Lu Li, Yu Qin, Zhuyi Wang for suggestions.

\section*{Contributions}
\begin{itemize}
\item SZ conceived the project;
\item SZ, PG, and YB directed the project;
\item All authors developed the conceptual and theoretical framework; 
\item TZ, AW, SP, SS developed the economic simulator and implemented the reinforcement learning platform;
\item SS developed the evaluation pipeline. 
\item SS developed the GPU version of the simulation.
\item TZ scraped data;
\item TZ, SP performed calibration;
\item TZ, AW, SP, SS performed experiments;
\item TZ, SP, AW developed the tutorial notebook;
\item YZ advised the work;
\item All authors drafted, discussed, and commented on the manuscript.
\item All authors reviewed the code.
\end{itemize}
\section{The Activity Component: Climate, Economics, Trade, and Tariffs}
\label{sec:dynamics_appendix}

\subsection{Climate and Economic Dynamics}
\label{sec:climateecon_appendix}
We now describe the RICE-N dynamics developed from DICE and RICE models by \cite{RePEc:spr:climat:v:148:y:2018:i:4:d:10.1007_s10584-018-2218-y, kellett2019feedback} that govern the evolution of the world state from time $\idstep$ to $\idstep+1$ for the different regions. 
Note that variables without an agent index are global quantities.


%

%



\begin{figure}
    \centering
    \includegraphics[width=8cm]{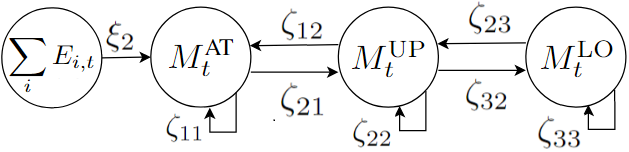}
    \caption{
    \textbf{The three-reservoir carbon mass model.}
    }
    \label{fig:three-reservoir}
\end{figure}

\paragraph{Carbon mass.}
The total carbon mass in the climate system is given by:
\eq{
\xcarbonmass_{\idstep+1} &=\Phi_{M} \xcarbonmass_{\idstep}+B_{M} \sum_{i}\xemission_{\idxi, \idstep}, \\
\xemission_{\idxi, \idstep} &= \xemission^{\textrm{Land}}_{\idstep}+ \xcarbonintensity_{\idxi, \idstep} \xtechnology_{\idxi, \idstep} (1-\mu_{\idxi, \idstep}) \xproduction_{\idxi, \idstep} \label{eq:sigma-t}, \\
\xcarbonmass_{\idstep} &\doteq\left[\begin{array}{lll}\xcarbonmass^{\textrm{AT}}_{\idstep} & \xcarbonmass^{\mathrm{UP}}_{\idstep} & \xcarbonmass^{\mathrm{LO}}_{\idstep}\end{array}\right]^{\top} \in \mathbb{R}^{3}, \\
\Phi_{M} &\doteq\left[\begin{array}{ccc}\zeta_{11} & \zeta_{12} & 0 \\
\zeta_{21} & \zeta_{22} & \zeta_{23} \\ 0 & \zeta_{32} & \zeta_{33}\end{array}\right], \\
B_{M} &\doteq\left[\begin{array}{c}\xi_{2} \\ 0 \\ 0\end{array}\right].
}
This describes a three-reservoir model of the global carbon cycle, in which $M_{AT}$ describes the average mass of carbon in the atmosphere, $M_{UP}$ is the average mass of carbon in the upper ocean, and $M_{LO}$ the average mass of carbon in the deep or lower ocean, see Figure \ref{fig:three-reservoir}. 
$\Phi_{M}$ is the Markov transition matrix describing how carbon transfer between different reservoirs. 
$B_{M}$ describes how the weight of carbon emission affects the carbon accumulation in the reservoirs.

\paragraph{Global temperature.}
Ultimately, increasing carbon mass leads to rising temperatures:
\eq{
    \xtemperature_{\idstep+1} &=\Phi_{T} \xtemperature_{\idstep}+B_{T}\xforcing_{\idstep}, \\
    \xtemperature_{t} &\doteq\left[\begin{array}{lll}\xtemperature^{\textrm{AT}}_{t}& \xtemperature^{\textrm{LO}}_{t}\end{array}\right]^{\top} \in \mathbb{R}^{2}, \\
    \xforcing_{\idstep} &= F_{2 \times} \log _{2}\left(\frac{\xcarbonmass^{\textrm{AT}}_\idstep}{\xcarbonmass^{\textrm{AT}, 1750}}\right), \\
    \Phi_{T} &\doteq\left[\begin{array}{ll}\phi_{11} & \phi_{12} \\ \phi_{21} & \phi_{22}\end{array}\right], \\
    B_{T} &\doteq\left[\begin{array}{l}\xi_{1} \\ 0\end{array}\right].
}

\begin{figure}
    \centering
    \includegraphics[width=4cm]{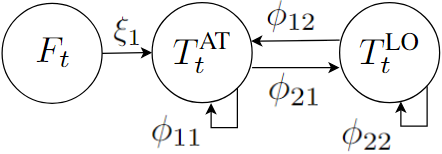}
    \caption{\textbf{The two-reservoir temperature model.}}
    \label{fig:energy-model}
\end{figure}

Similar to the carbon mass dynamic, there are two layers in the energy balance model, see Figure \ref{fig:energy-model}.
$T_{AT}$ is the combined average temperature in atmosphere, land surface, and upper ocean (simply referred to as the ``atmospheric layer'' hereafter). 
$T_{LO}$ is the temperature in the lower ocean. 
$\Phi_{T}$ is the Markov transition matrix describing how heat transfers between different layers. 
$B_{T}$ describes how carbon mass contributes to the temperature increases. 


 
\paragraph{Output production.}
The production in a region is given by the \emph{total factor productivity (TFP)} \cite{Comin2010} formula:
\eq{ \tag{\ref{eq:rice:production}}
\xproduction_{\idxi, \idstep} = \xtechnology_{\idxi, \idstep}\xcapital_{\idxi, \idstep}^{\gamma}\xpopulation_{\idxi, \idstep}^{1-\gamma}.
}
Production depends on three factors: total factor productivity (``technology'') $\xtechnology_{\idstep}$, capital $\xcapital_{\idstep}$, and labor $\xpopulation_{\idstep}$. 
This production function is common in the economic literature and used in the DICE/RICE models. 
The capital elasticity $\gamma \in [0,1]$ explains the different levels of contribution of capital and labor.

\paragraph{Population.}
The number of people in a region, denoted $\xpopulation_{\idstep}$ grows as:
\begin{equation} \tag{\ref{eq:rice:population}}
\xpopulation_{\idxi, \idstep+1} =\xpopulation_{\idxi, \idstep}\brck{ \frac{1+\xpopulation_{a; \idxi}}{1+\xpopulation_{\idxi, \idstep}} }^{l_{g; \idxi}}.
\end{equation}

There are two parameters $L_{A;\idxi}$ and $l_{g; \idxi}$. $L_{a; \idxi}$ represents the convergence population of region $\idxi$ and $l_{g; \idxi}$ shows how fast the population $L_{\idxi, \idstep}$ converge to $L_{a; \idxi}$. 
Please refer to the Appendix \ref{sec:model-calibration} for a more detailed analysis and the calibration procedure.

\paragraph{Level of technology.}
The technology factor $A_{\idstep}$ describes how efficient production is, i.e., how many units of output a region achieves given fixed capital and labor:
%
\eq{ \tag{\ref{eq:rice:technology}}
\xtechnology_{\idxi, \idstep+1} &= (e^\eta + g_{A;\idxi} e^{-\delta_{A;\idxi} \Delta(\idstep-1)} ) \xtechnology_{\idxi, \idstep}.
}

Here, $\eta$ represents the long-term growth of economics which is usually larger than 0,
$g_A$ represents the short-term part of economics growth,
and $\delta_A$ represents the speed of decay of short-term growth factor.
$\Delta$ is the time difference between steps. 
We use $\eta = 0.33\%$ as in \cite{RePEc:spr:climat:v:148:y:2018:i:4:d:10.1007_s10584-018-2218-y}.

\paragraph{Capital.}
The amount of capital evolves as:
\eq{ 
\Phi_{K} &\doteq \left(1-\delta_{K}\right)^{\Delta}, \\
\xcapital_{\idxi, \idstep+1} &=\Phi_{K, \idxi} \xcapital_{\idxi, \idstep}+\Delta\brck{1-a_{1} \xtemperature^{\textrm{AT}}_{\idstep} - a_{2} \brck{\xtemperature^{\textrm{AT}}_{\idstep}}^{2} } \\
&\times \left(1-\theta_{1; \idxi,\idstep} \mu_{\idxi, \idstep}^{\theta_{2}}\right) \xproduction_{\idxi, \idstep} \xsavings_{\idxi, \idstep}.
}
The evolution of the capital comes from two parts. 
The first part is capital inherited from the previous period with depreciation. 
In the second part, $\xsavings_{\idstep}$ is a control variable which represents the investment/savings rate (as a fraction of production). 
That is, as a base amount, the economy invests/saves a total of $\xproduction_{\idxi,\idstep} \xsavings_{\idxi, \idstep}$ 
which yields new capital.
This base amount is further modified by 2 multipliers: the damage function and mitigation/abatement costs, which are discussed below.


\paragraph{Damage function.}
The climate damage function represents the economic damage due to climate change, e.g., increases in the atmosphere temperature $\xtemperature_{\idstep}^{\textrm{AT}}$. 
That is, in Equation \ref{eq:rice:capital}, the fraction of new capital is modified by the damage function
\eq{\label{eq:damage-function}
1-a_{1} \xtemperature^{\textrm{AT}}_{\idstep} - a_{2} \brck{\xtemperature^{\textrm{AT}}_{\idstep}}^{2},
} 
following \cite{nordhaus2015climate}.
That is, higher temperatures lead to less new capital.
Similarly, $1-\theta_{1;\idxi, \idstep} \mu_{\idstep}^{\theta_{2}}$ is the fraction of new capital after taking into account carbon emission mitigation.
Mitigating carbon emissions more (higher $\mu_{\idstep}$) means (dirty) production needs to be lowered, hence yields less new capital.

\paragraph{Mitigation (abatement) cost.}
Following \cite{kellett2019feedback}, for a mitigation rate $ \mu_{\idxi, \idstep}$, the mitigation cost is 
\eq{\label{eq:mitigation-cost}
\theta_{1; \idxi,\idstep} \mu_{\idxi, \idstep}^{\theta_{2}}\xproduction_{\idxi,\idstep} \xsavings_{\idxi, \idstep},} 
where $\theta_{1; \idxi, \idstep}$ is given by Equation \ref{eq:rice:cost_of_mitigation_efforts}.
This represents the loss in capital growth due to a fraction of production being used for mitigation.
%

\paragraph{Carbon intensity of economic activity.}
A critical part of the model is the interaction between the climate and economic parts. 
Specifically, the RICE model describes how production leads to carbon emissions:
\eq{
\xemission^{\textrm{Land}}_{\idstep}&=\xemission_{L0} \cdot\left(1-\delta_{EL}\right)^{t-1}, \\
\xemission_{\idxi, \idstep} &= \xemission^{\textrm{Land}}_{\idstep}+ \xcarbonintensity_{\idxi, \idstep} \xtechnology_{\idxi, \idstep} (1-\mu_{\idxi, \idstep}) \xproduction_{\idxi, \idstep} \\
\xcarbonintensity_{\idxi, \idstep+1} &=\xcarbonintensity_{\idxi, \idstep} e^ {-g_{\xcarbonintensity; \idxi}\brck{1-\delta_{\xcarbonintensity; \idxi}}^{\Delta(\idstep-1)} \Delta}.
}
Here $\xemission^{\textrm{land}}_{\idstep}$ is the carbon emission due to (changes in) land use, 
$\xemission_{L0}$ is the carbon emission in the base year, 
and $\delta_{EL}$ is the speed of decrease of changes in land use.
The rates $0 < \delta_{EL} < 1$, $0 <\delta_{L0} < 1$ are free parameters. 
Due to a lack of data, $\xemission^{\textrm{land}}_{\idstep}$ is set to be the same for each region.

$\xemission_{\idxi, \idstep}$ is the total carbon emission, $\xemission_{t}^{\textrm{land}}$ is emission from natural sources, while $\xemission_{\idxi, \idstep} - \xemission_{t}^{\textrm{land}}$ is emission caused by economic activity.
$\xcarbonintensity_{\idxi, \idstep}\xtechnology_{\idxi,\idstep}$ is the effective carbon intensity of economic activity: a higher technology factor lead to higher emissions, but can be modulated by lower $\xcarbonintensity$ (which can be thought of as the degree of ``clean'' production).
$\mu_{idxi, \idstep} \in [0, 1]$ is a control variable called the abatement (ratio), which represents the proportion of the economics contributing to reducing carbon emission. 
Furthermore, we have 2 parameters $g_{\xcarbonintensity}$ and $\delta_{\xcarbonintensity}$ that are fitted to data. 
$g_{\xcarbonintensity}$ is the rate of decrease in carbon emissions.



\subsection{Trade}
\label{sec:trade_appendix}

We now describe the international trade dynamics and the resulting regional consumption and utilities.
Regions trade by exporting their own consumption goods and importing other regions' consumption goods at a fixed unit price\footnote{More generally, prices should be dynamic, but the current implementation does not support this.}.


\paragraph{Agent actions.}
Each region $\idxi$ at time $t$ must first specify a desired basket of consumption goods $ \bm{\ximportbid_{\idxi,t}} = [\ximportbid_{\idxi,1,t}, ..., \ximportbid_{\idxi, \idxk,t}]$ that they are willing to import from the other regions.
These desired imports form a matrix of \textit{bids} $\ximportbidmatrix_{\idstep}$ such that the import bid by region $\idxi$ for goods from $\idxj$ at time $\idstep$ is $\ximportbid_{\idxi,\idxj,\idstep} \geq 0$, i.e., the amount of goods region $\idxi$ is willing to import from region $\idxj$ at time $\idstep$ is $\ximportbid_{\idxi,\idxj,\idstep}$.

Regions also set an upper bound $\xexportlimit_{\idxi, \idstep} \in [0,1]$ on the proportion of their own consumption goods that they are willing to export.

%

\paragraph{Tariffs.}
Regions can also choose to impose import tariffs on other regions.
We denote an import tariff imposed by region $\idxi$ on a region $\idxj$ by $\xtariff_{\idxi,\idxj,\idstep} \in [0,1]$.
If region $\idxi$ imposes an import tariff $\xtariff_{\idxi,\idxj,\idstep} \in [0,1]$ on region $\idxj$, region $\idxi$ consumes 
\begin{equation} \label{eq:trade:tariff_goods}
    \xconsumption_{\idxi,\idxj,\idstep} = \xgoodstraded_{\idxi,\idxj,\idstep} (1 - \xtariff_{\idxi,\idxj,\idstep}),
\end{equation}


and $\tau_{\idxi, \idxj, \idstep}x_{\idxi,\idxj,\idstep}$ is added to a reserve fund specific to that region.
\paragraph{Consumption.}

Consumption of domestic goods $\xconsumption_{i,i,t}$ is determined according to gross output, the savings rate and exports:

\begin{equation}
    \label{eq:consumption_domestic}
    \xconsumption_{i,i,t} = (1 - \xsavings_{i,t})\xgrossoutput_{i,t} - \sum_{\idxj \neq \idxi} \xgoodstraded_{\idxj,\idxi,t}.
\end{equation}

The aggregated consumption $C_{\idxi, \idstep}$ at time $t$ for region $\idxi$ is given by the Armington elasticity model~\cite{lessmann2009effects}) as follows:
\begin{equation} \label{eq:agg_consumption}
    C_{\idxi, \idstep} = \brck{  \psi^{dom}(\xconsumption_{\idxi, \idxi, \idstep})^\lambda +  \sum_{j\neq i} \psi^{for} (\xconsumption_{\idxi, \idxj, \idstep})^\lambda  }^{\fr{1}{\lambda}}.
\end{equation}





\section{\SimulationName{} dynamics}
\label{sec:rice-dynamics-details}

At a high-level, Equations \ref{eq:rice:temperature} and \ref{eq:rice:carbon} capture climate dynamics (temperature and carbon mass), while Equations \ref{eq:rice:capital}, \ref{eq:rice:population}, \ref{eq:rice:technology}, and \ref{eq:rice:production} capture economic dynamics. 
Finally, Equation \ref{eq:rice:carbon-intensity} captures the carbon-intensity of production, providing a key link between the climate and economic sectors.

\begin{algorithm*}
\small
\caption{
Activity Component (implemented by \texttt{Climate\_and\_economy\_simulation\_step()})
Note that we only list input state variables and omit model parameters.
}\label{alg:cap}
\begin{algorithmic}
\Require exogenous emissions, land emissions, intensity, production factor, labor, capital, \\ previous global temperature, previous government balance
\Require actions : mitigation rates, saving rates, tariffs, export rate limit, desired imports
\For{each region}
    \State $\text{mitigation cost} \gets f(\text{intensity})$ \Comment{Equation \ref{eq:rice:cost_of_mitigation_efforts}}
    \State $\text{damages} \gets f(\text{previous global temperature})$ \Comment{Equation \ref{eq:damage-function}}
    \State $\text{abatement cost} \gets f(\text{mitigation rate, mitigation cost})$ \Comment{Equation \ref{eq:mitigation-cost}}
    \State
    \State $\text{production} \gets f(\text{production factor, capital, labor})$ \Comment{Equation \ref{eq:rice:production}}
    \State $\text{gross output} \gets f(\text{damages, abatement cost, production})$ \Comment{Equation \ref{eq:rice:production}}
    \State $\text{government balance} \gets f(\text{interest rate},\text{previous government balance})$ 
    \State $\text{investment} \gets f(\text{saving rate, gross output})$ \Comment{Using Equation \ref{eq:rice:capital}}
    \State
    \State $\text{scaled imports} \gets f(\text{gross output, desired imports})$  \Comment{Equation \ref{eq:trade:norm_1}}
    \State $\text{debt ratio} \gets f(\text{previous government balance})$ \Comment{Equation \ref{eq:trade:debt_ratio}}
    \State scaled imports $\gets f(\text{scaled imports, debt ratio})$ \Comment{Equation \ref{eq:trade:norm_2}}
\EndFor
\For {each region}
    \State $\text{max potential exports} \gets f(\text{gross output, investment, export rate limit})$ \Comment{Equation \ref{eq:trade:export_max}}
    \State $\text{Scaled imports} \gets f(\text{scaled imports, max potential exports})$ \Comment{Equation \ref{eq:trade:norm_3}}
\EndFor
\For {each region}
    \State tariff-ed imports, tariff revenue $\gets f(\text{scaled imports, tariffs})$  \Comment{Equation \ref{eq:trade:tariff_goods}}
    \State domestic consumption $\gets f(\text{savings, gross output, scaled imports)}$ \Comment{Equation \ref{eq:consumption_domestic}}
    \State aggregate consumption $\gets f(\text{domestic consumption, tariff-ed imports})$ \Comment{Equation \ref{eq:agg_consumption}}
    \State utility $\gets f(\text{labor, aggregate consumption})$ \Comment{Equation \ref{eq:utility}}
    \State government balance $\gets f(\text{imports, exports})$ \Comment{Equation \ref{eq:trade:balance}}
\EndFor
    \State temperature $\gets f(\text{previous temperature, previous carbon mass, exogenous emissions})$
    \State carbon mass $\gets f(\text{previous carbon mass, intensity, mitigation rate, production, land emissions})$
\For {each region}
    \State capital $\gets f(\text{capital, investment})$ \Comment{Equation \ref{eq:rice:capital}}
    \State labor $\gets f(\text{labor})$ \Comment{Equation \ref{eq:rice:population}}
    \State production factor $\gets f(\text{capital})$ \Comment{Equation \ref{eq:rice:technology}}
    \State carbon intensity $\gets f(\text{carbon intensity})$ \Comment{Equation \ref{eq:rice:carbon-intensity}}
\EndFor
    
\end{algorithmic}
\end{algorithm*}
\section{Creating a 27-Region Simulation}
\label{sec:merging-splitting-regions}
We feature $\numRegions = 27$ fictitious regions in our public simulation. 
These are inspired by merging and splitting real-world countries, but are not exactly the same as real-world regions.

We used real data from the World Bank API \cite{wbapi}, e.g., GDP, capital stock, population, and $\text{CO}_{2}$-quivalent ($\text{CO}_{2}\text{eq}$) emissions.
Furthermore, the World Bank groups countries into regions, including Sub-Saharan Africa, South Asia, North America, the Middle East and North Africa, Latin America and the Caribbean, Europe and Central Asia, East Asia and Pacific. 
In each region, the different countries (or sub-regions) are classified into 4 income groups: high income, upper middle income, lower middle income, and low income.

\paragraph{Merging regions.}
We assume the GDP, capital stock, and population for the regions are additive. 
We also assume the gross $\text{CO}_{2}\text{eq}$ emissions across the regions are additive. 
Thus, we have
\begin{align}
    K_m &= \sum_{i} K_i, \\ 
    L_m &= \sum_{i} L_i, \\
    Y_m &= \sum_{i} Y_i, \quad \text{where} \quad Y_i := A_i K_i^{\gamma} L_i^{1-\gamma}, \\
    A_m &= \frac{Y_m}{K_m^{\gamma} L_m^{1-\gamma}}, \\
    \sigma_{m} &= \frac{\sum_{i} \sigma_{i} Y_{i}}{Y_m}.
\end{align}
Note that the production function is not scale-invariant:
\eq{ 
Y_t &= (A_t K_t )^{\gamma}(A_t \xpopulation_{\idstep})^{1-\gamma} \\
c\cdot Y_t &= (c\cdot A_t K_t)^{\gamma}(c\cdot A_t \xpopulation_{\idstep})^{1-\gamma} \\
&\neq (c\cdot A_t )(c\cdot K_t )^{\gamma}(c\cdot \xpopulation_{\idstep})^{1-\gamma}, \quad \forall c > 0. 
}
Hence, one cannot get the technology after merging multiple regions by simply adding the individual technology levels. Rather, the combined technology factor is imputed from the combined productions, labor, and capital.
\begin{figure*}
\begin{align}
\xtemperature_{\idstep+1} &=\Phi_{T} \xtemperature_{\idstep}+B_{T}\left(F_{2 \times} \log _{2}\left(\frac{\xcarbonmass^{\textrm{AT}}_{\idstep}}{\xcarbonmass^{\textrm{AT}, 1750}}\right)+F^{\textrm{EX}}_{\idstep}\right), \label{eq:rice:temperature} \\ 
\xcarbonmass_{\idstep+1} &=\Phi_{M} \xcarbonmass_{\idstep}+B_{M}\brck{
\sum_\idxi \xcarbonintensity_{\idxi, \idstep}(1-\mu_{\idxi, \idstep}) \xproduction_{\idxi, \idstep} + \xemission^{\textrm{Land}}_{\idstep} }, \label{eq:rice:carbon} \\ 
\xcapital_{\idxi, \idstep+1} &=\Phi_{K, \idxi} \xcapital_{\idxi, \idstep}+\Delta\brck{1-a_{1} \xtemperature^{\textrm{AT}}_{\idstep} - a_{2} \brck{\xtemperature^{\textrm{AT}}_{\idstep}}^{2} } \left(1-\theta_{1; \idxi,\idstep} \mu_{\idxi, \idstep}^{\theta_{2}}\right) \xproduction_{\idxi, \idstep} \xsavings_{\idxi, \idstep}, \label{eq:rice:capital}\\ 
\theta_{1; \idxi, \idstep} &= \frac{p_{b}}{1000 \cdot \theta_{2}}\left(1-\delta_{p b}\right)^{\idstep-1} \cdot \xcarbonintensity_{\idxi, \idstep}, \label{eq:rice:cost_of_mitigation_efforts}\\
\xpopulation_{\idxi, \idstep+1} &=\xpopulation_{\idxi, \idstep}\brck{ \frac{1+\xpopulation_{a; \idxi}}{1+\xpopulation_{\idxi, \idstep}} }^{l_{g; \idxi}}, \label{eq:rice:population}\\ 
\xtechnology_{\idxi, \idstep+1} &= (e^\eta + g_{A;\idxi} e^{-\delta_{A;\idxi} \Delta(\idstep-1)} ) \xtechnology_{\idxi, \idstep}, \label{eq:rice:technology} \\ 
\xproduction_{\idxi, \idstep} &= \xtechnology_{\idxi, \idstep}\xcapital_{\idxi, \idstep}^{\gamma}\xpopulation_{\idxi, \idstep}^{1-\gamma}  \label{eq:rice:production}, \\
\xcarbonintensity_{\idxi, \idstep+1} &=\xcarbonintensity_{\idxi, \idstep} e^ {-g_{\xcarbonintensity; \idxi}\brck{1-\delta_{\xcarbonintensity; \idxi}}^{\Delta(\idstep-1)} \Delta}. \label{eq:rice:carbon-intensity}
\end{align}
\end{figure*}

\paragraph{Splitting large regions.}
To avoid huge economies that dominate the fictitious world, we split large economies into pieces based on predetermined fractions $c_i$ and random sampled $A_i$:
\begin{align}
    \sum_{i} c_i &= 1, \\ 
    L_i &= c_i L_m,  \\
    Y_i &= c_i Y_m, \\
    K_i &= \frac{Y_i}{A_i L_i^{1-\gamma}}, \\
    \sigma_{i} &= \sigma_{m}.
\end{align}

\section{Model Calibration}
\label{sec:model-calibration}

The structural parameters of the \SimulationName{} simulation were calibrated to meet the following objectives:

\begin{enumerate}
    \item Temperatures match the real data in different versions of \SimulationName{} with 1 region, 27 regions, and 189 regions (all raw regions), under 0\% and 100\% mitigation. To make the computational cost more affordable, we use the 27-region version for the competition.
    \item The optimistic-pessimistic temperature outcomes fit the projects of Shared Socioeconomic Pathways in the IPCC Sixth Assessment Report \cite{portner2022climate} (2 - 5 deg Celsius increase in the year of 2100). Each region optimizes the target without negotiation and direct cooperation in the pessimistic case. In the optimistic case, regions negotiate with each other using the baseline bilateral negotiation protocol. Please also notice that, in the extreme pessimistic case that regions ignore the climate change at all and always choose 0\% mitigation and 100\% savings, the temperature leads to approximately 7 deg Celsius increase in the year of 2100.
\end{enumerate}

The parameters that we estimated and the corresponding estimation methods are listed below:
\begin{itemize}
    \item The dynamic parameters for total factor productivity $A$: $g_A$ and $\delta_A$.
    \item The capital $\xcapital$: for the regions whose capital data is not available, we use a KNN regressor~\cite{sklearn_api} to estimate it.
    \item The dynamic parameters for population $\xpopulation$: $l_g$; similarly, for the regions whose convergence population data is not available, we use a KNN regressor to estimate it.
    \item The initial carbon intensity $\xcarbonintensity_0$: for the regions whose capital data is not available, we use a KNN regressor to estimate it.
    \item KNN regressor: Because all regions have GDP and population data, we use them as features. 
    For each region that lacks emission data and capital data, we find the nearest 5 neighbors according to its GDP and population. 
    We use the average of the 5 neighbors' emission data and capital data as the estimated values.
\end{itemize}

\subsection{Population dynamic calibration}
Denoting $\xpopulation_{\infty; \idxi}:=\lim_{\idstep\rightarrow \infty} \xpopulation_{\idxi, \idstep}$, in the limit $t \rightarrow \infty$ we have:
\eq{
\xpopulation_{\infty; \idxi} &= \xpopulation_{\infty; \idxi} \brck{\frac{1+\xpopulation_{a,\idxi}}{1+\xpopulation_{\infty; \idxi}}}^{l_{g; \idxi}}, \\
 1 &= \brck{\frac{1+\xpopulation_{a; \idxi}}{1+\xpopulation_{\infty}}}^{l_{g; \idxi}}.
}
As long as $l_{g; \idxi}$ is not zero, $\xpopulation_{\infty; \idxi} = \xpopulation_{a; \idxi}$. Thus, $L_{a; \idxi}$ is the long-term population size and a free parameter that is fitted to data. 
Assuming $\{\xpopulation_{\idxi, \idstep}\}_{t=1,2,\dots}$ is monotonically increasing or decreasing, the absolute value of $l_{g; \idxi}$ represents how fast it converges to $\xpopulation_{a; \idxi}$.
The closer $\xpopulation_{\idxi, \idstep}$ is to monotonically increasing or monotonically decreasing in the real data, the easier it is to fit $l_{g; \idxi}$ and $ \xpopulation_{a; \idxi}$.

To fit the population parameters, we take logs on both sides of Equation \ref{eq:rice:population}: 
\eq{
& \log{\xpopulation_{\idxi, \idstep+1}} = \notag\\ 
& \log{\xpopulation_{\idxi, \idstep}} + l_{g; \idxi} (\log{(1+L_{a; \idxi})} - \log{(1+\xpopulation_{\idxi, \idstep+1})}),
}
where $\log{\xpopulation_{\idxi, \idstep+1}} - \log{\xpopulation_{\idxi, \idstep}}$ and $\log{(1+\xpopulation_{\idxi, \idstep})}$ are given by the data.
$\log{(1+\xpopulation_{\idxi, \idstep})}$ and $l_{g; \idxi}$ can then be estimated by linear regression.

\subsection{Technology dynamic calibration}
We estimate both $g_A$ and $\delta_A$ from the existing data $\{\xtechnology_t\}_{i=1 \cdots n}$ by solving a regression problem:
\eq{
& g_{a; \idxi}^{*}, \delta_{a; \idxi}^{*} = \argmax_{g_{a; \idxi}, \delta_{A,\idxi}} \mathcal{L}_{\idxi, \idstep} \\ 
& \mathcal{L}_{\idxi, \idstep} = ||A_{\idxi, \idstep} - (\exp\eta+g_{A,\idxi} \exp \left(-\delta_{A,\idxi} \Delta(t-1)\right))\xtechnology_{\idxi,\idstep}||^2.
}
This can be solved by numerical optimization algorithms, e.g., as provided in SciPy \cite{2020SciPy-NMeth}.

Because the emissions data from the World Bank API do not fit the form of the $\xcarbonintensity$ dynamic as assumed by DICE2016, use the DICE2016 parameter values for $g_{\xcarbonintensity}$ and $\delta_{\xcarbonintensity}$.

\begin{table*}[t]
\centering
\caption{Calibrated parameters for 27 regions}
\begin{tabular}{ccccccccc}
Region ID & $A_0$  & $K_0$  & $L_0$   & $L_a$    & $delta_A$ & $g_A$ & $l_g$  & $sigma_0$ \\
\hline \\
1         & 1.872  & 0.239  & 476.878 & 669.594  & 0.139     & 0.122 & 0.034  & 0.456     \\
2         & 8.405  & 3.304  & 68.395  & 93.497   & 0.188     & 0.103 & 0.058  & 0.529     \\
3         & 3.558  & 0.109  & 64.122  & 135.074  & 0.161     & 0.127 & 0.026  & 0.816     \\
4         & 1.927  & 1.424  & 284.699 & 465.308  & 0.244     & 0.134 & 0.024  & 1.221     \\
5         & 8.111  & 0.268  & 28.141  & 23.574   & 0.163     & 0.106 & -0.057 & 0.290     \\
6         & 4.217  & 3.184  & 548.754 & 560.054  & 0.170     & 0.095 & 0.080  & 0.302     \\
7         & 2.491  & 0.044  & 46.489  & 59.988   & 0.058     & 0.049 & 0.037  & 0.420     \\
8         & 2.525  & 1.080  & 69.194  & 100.016  & 0.346     & 0.079 & 0.029  & 1.010     \\
9         & 2.460  & 0.184  & 513.737 & 1867.771 & 1.839     & 0.462 & 0.017  & 0.310     \\
10        & 12.158 & 2.642  & 38.101  & 56.990   & 0.131     & 0.063 & 0.020  & 0.350     \\
11        & 0.993  & 0.160  & 522.482 & 1830.325 & 0.086     & 0.065 & 0.019  & 0.235     \\
12        & 5.000  & 2.289  & 165.293 & 230.191  & 0.183     & 0.071 & 0.027  & 0.419     \\
13        & 29.854 & 2.020  & 165.751 & 216.927  & 0.088     & 0.075 & -0.002 & 0.254     \\
14        & 23.315 & 3.039  & 109.395 & 143.172  & 0.088     & 0.075 & -0.002 & 0.254     \\
15        & 29.854 & 0.687  & 56.355  & 73.755   & 0.088     & 0.075 & -0.002 & 0.254     \\
16        & 10.922 & 0.606  & 705.465 & 532.497  & 0.096     & 0.168 & -0.016 & 0.781     \\
17        & 9.634  & 0.608  & 465.607 & 351.448  & 0.096     & 0.168 & -0.016 & 0.781     \\
18        & 8.621  & 0.453  & 239.858 & 181.049  & 0.096     & 0.168 & -0.016 & 0.781     \\
19        & 3.190  & 0.129  & 690.002 & 723.513  & 0.054     & 0.068 & -0.013 & 0.949     \\
20        & 2.034  & 0.381  & 455.401 & 477.518  & 0.054     & 0.068 & -0.013 & 0.949     \\
21        & 13.220 & 16.295 & 502.410 & 445.861  & 0.252     & 0.074 & -0.033 & 0.170     \\
22        & 3.190  & 0.044  & 234.601 & 245.994  & 0.054     & 0.068 & -0.013 & 0.949     \\
23        & 6.387  & 1.094  & 317.880 & 287.533  & 0.194     & 0.237 & -0.053 & 0.840     \\
24        & 2.481  & 0.090  & 94.484  & 102.997  & 0.203     & 0.201 & 0.037  & 1.665     \\
25        & 10.853 & 17.554 & 222.891 & 168.351  & 0.005     & 0.000 & -0.012 & 0.285     \\
26        & 4.135  & 1.002  & 103.294 & 87.418   & 0.158     & 0.123 & -0.063 & 0.601     \\
27        & 2.716  & 1.034  & 573.818 & 681.210  & 0.097     & 0.101 & 0.043  & 0.638    
\end{tabular}
\label{tab:params}
\end{table*}
\section{Implementation details}
\label{sec:engineering-details}

\paragraph{Consistency checks.}
On the other hand, for the purpose of the competition, we fix the actions in the activity stage and the corresponding climate and economic dynamics.
Thus, certain parts of the simulation should not be modified by competitors, e.g., the core climate and economic parameters and equation, such as the ones which affect carbon emissions and productivity.
Changing these components may change the problem and prevent fair comparisons across submitted solutions.
As such, our evaluation protocol and implementation of \SimulationName{} include consistency checks that test submitted solutions; each submission requiring submitting the full code of the (modified) simulation and agent models.
The activity stage is designed to follow 
the well established climate-economic dynamics in the RICE model 
augmented with trades and tariffs to include an essential influence in international negotiations.
This way the simulation follows a simple model but is rich enough to implement  interesting negotiation protocols.

Further, our implementation is set up to model each region as an RL agent that optimizes its discounted long-term rewards.
Although other agent policy models are possible, we require that the competitors clarify how their proposed agent policy models also aim to optimize for the same goal, namely, the personal discounted long term reward for each agent.
This ensures that the agents behave greedily and any improvements in global metrics are strictly a result of the implemented negotiation protocols.
See Section \ref{sec:AgentModels} for details on the agent model.

\paragraph{CPU vs GPU simulations.}
We provide both CPU and GPU versions of the base \SimulationName{}, but evaluate all submissions using the CPU version to reduce the cost of evaluation. 
The CPU and GPU implementations of the base version of \SimulationName{} yield consistent results. 
The GPU version of the simulation is meant to accelerate training. 
\section{$\text{CO}_{2}$ Emission}
\label{sec:CO2-emissions}
Participants who use Google Colab have a carbon efficiency of 0.37 kg $\text{CO}_{2}\text{eq}$/kWh. This was measured on 0.35 hours of computation on a Tesla T4 (300W TDP) node.

The emissions produced by going through the tutorial and training a baseline model are estimated to be 0.01 kg $\text{CO}_{2}\text{eq}$ of which 100\% are offset by the cloud provider.

Estimates were obtained using the Machine Learning Impact calculator~\cite{lacoste2019quantifying} available at \url{https://mlco2.github.io/impact#compute}.
We encourage participants to use tools such as CodeCarbon~\cite{codecarbon} (available at \url{https://codecarbon.io}) as illustrated in the tutorial notebook to track their carbon emissions and report it as part of their publications. 
\begin{figure*}[t]
     \renewcommand*\thesubfigure{\roman{subfigure}} 
     \centering
     \begin{subfigure}[b]{0.4\textwidth}
         \centering
         \includegraphics[width=\textwidth]{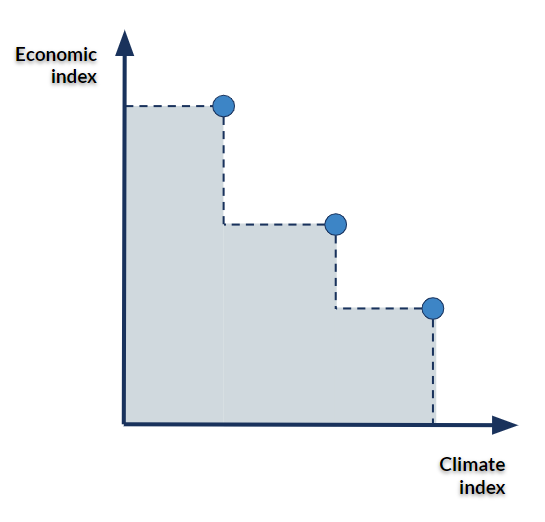}
         \caption{}
         \label{pareto-single}
     \end{subfigure}
     \begin{subfigure}[b]{0.405\textwidth}
         \centering
         \includegraphics[width=\textwidth]{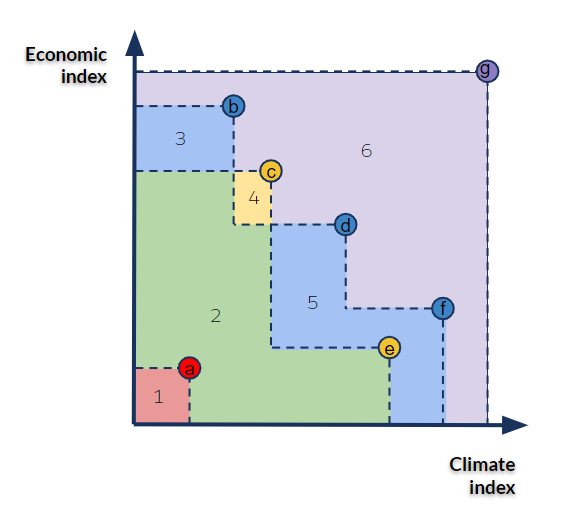}
         \caption{}
         \label{fig:three sin x}
     \end{subfigure}
        \caption{
        \textbf{A visualization of how teams are ranked using the hypervolume indicator~\cite{zitzler2003performance}.}
        (i) A single team's set of submissions is represented by a set of points of the same color.
        The hypervolume indicator is the area of a set of non-dominated policies~\cite{van2014multi} in the 2-dimensional $(\climateindex, \economicindex)$-space, as represented by the shaded area.
        (ii) Multiple sets of solutions (\{a\}, \{b,d,f\}, \{c,e\}, \{g\})proposed by various teams.
        Teams are ranked by the value of their hypervolume indicators: a higher value is a better score.
        A first team submits point (a), so their set of solutions contains one point.
        The hypervolume indicator of this solution corresponds to area 1.
        A second team submits points (c) and (e).
        Their score corresponds to the sum of areas 2 and 4.
        A third team submits points (b), (d) and (f), and so their score is the sum of areas 2, 3 and 5.
        A final team submits a single point (g).
        The hypervolume indicator of this point corresponds to the sum of all areas 1 to 6.
        As such, the team that submitted point (g) is ranked highest, the team that submitted points (b), (d) and (f) is second, the team that submitted points (c) and (e) is third, and the team that submitted point (a) is last. 
        }
        \label{fig:hypervolume}
\end{figure*}

\section{Submission and evaluation details}
\label{sec:submission-evaluation-details}

\paragraph{Structure of a submission.}
Submissions consist of the full simulation code, agent behavioral models, and other components that may be needed to generate full simulation roll-outs.

If participants use the GPU version of the simulation, they should also provide a CPU version and ensure that any models trained on the GPU version and any customization yield consistent results on a CPU.

\paragraph{Evaluation metrics.}
Solutions are evaluated by computing the area under the empirical Pareto frontier produced by that solution. 
The area computation is explained in Figure \ref{fig:hypervolume}.

\section{WarpDrive Support}

Performing multi-agent RL is significantly faster using GPU acceleration. 
We provide a GPU version of the simulation and an end-to-end GPU-based RL training loop using WarpDrive.
For details, see the instruction on the Github repo \url{https://github.com/mila-iqia/climate-cooperation-competition}.
\section{Trade constraints}

To ensure that total imports and total exports match, three constraints are enforced on regions' trade flows.

\begin{enumerate}[leftmargin=*]
    \item For each region $\idxi$, if the region's total desired imports from other regions exceed its own gross output, then the imports are scaled to sum up to the region's gross output.
    We enforce the constraint that $\sum_{\idxi \neq \idxj} \ximportbid_{\idxi,\idxj,\idstep} \leq \xgrossoutput_{\idxi, \idstep}$, which is to say that a region may not import more goods than its current gross output capacity.
    This constraint helps the agents avoid insurmountable debt, thereby stabilizing trade balances over the entire time period while also easing learning.
    If a region's desired imports exceed its production capacity, then its import bids are scaled down to size : 
    \begin{equation}\label{eq:trade:norm_1}
        \ximportbid_{\idxi,\idxj,\idstep} \gets \ximportbid_{\idxi,\idxj,\idstep} \min\left\{1, \frac{\xgrossoutput_{\idxi, \idstep}}{\sum_{\idxi \neq \idxj} \ximportbid_{\idxi,\idxj,\idstep}} \right\}.
    \end{equation}

    \item Regions are allowed to carry a (positive or negative) trade balance $\xtradebalance_{i,t}$.
    At the start of each new time step, each region's trade balance, positive or negative, accumulates interest at a fixed rate of 10\%.
    Based on this balance, a region's debt-to-initial-capital ratio is determined and the imports are scaled according to this ratio:
    \begin{equation} \label{eq:trade:debt_ratio}
        \xdebtratio_{i,t} = 10 \frac{\xtradebalance_{i,t}}{K_0},
    \end{equation}
    \begin{equation} \label{eq:trade:norm_2}
        \ximportbid_{\idxi,\idxj,\idstep} \gets \ximportbid_{\idxi,\idxj,\idstep} (1 + \xdebtratio_{i,t}).
    \end{equation}

    \item If other regions' total desired imports from region $\idxi$ exceed region $\idxi$'s upper bound on exports $\xmaxexport_{i,t}$, then the bids for goods from region $\idxi$ are scaled proportionally to $\xmaxexport_{i,t}$.
    Otherwise, each region receives its full import bid from region $\idxi$.
    In other words, region $\idxi$ cannot export more goods at time $t$ than it could consume at time $t$, so other regions will import less from region $\idxi$.
    \begin{equation} \label{eq:trade:export_max} \xmaxexport_{i,t} = \min(\xexportlimit_{i,t}\xgrossoutput_{i,t}, \xgrossoutput_{i,t} - \xinvest_{i,t}),
    \end{equation}
    \begin{equation} \label{eq:trade:norm_3}
        \xgoodstraded_{\idxi,\idxj,\idstep} = \ximportbid_{\idxi,\idxj,\idstep} \min\left\{1, \frac{\min(\xmaxexport_{i,t})}{\sum_{\idxj \neq \idxi} \ximportbid_{\idxi,\idxj,\idstep}} \right\}.
    \end{equation}
\end{enumerate}

After all constraints have been applied, the trade balance for the next period is calculated:
\begin{equation} \label{eq:trade:balance}
    \xtradebalance_{\idxi, \idstep + 1} = \xtradebalance_{\idxi, \idstep} + \numStepPeriod\left(\sum_{\idxj \neq \idxi} \xgoodstraded_{\idxj, \idxi,t} - \sum_{\idxj \neq \idxi} \xgoodstraded_{\idxi, \idxj,t}\right).
\end{equation}

\section{Parameters and variables}
\label{sec:params-and-vars}

Tables \ref{tab:params}, \ref{tab:world_state_var}, \ref{tab:agent_action_var}, \ref{tab:agent_charact}, and \ref{tab:global_constants} list all (calibrated) parameters and variables.

\begin{table*}[t]
\centering
\small
\begin{tabular}{p{0.15\linewidth}p{0.17\linewidth}p{0.22\linewidth}p{0.35\linewidth}}
\textbf{Variable} & \textbf{Type} & \textbf{Symbol} & \textbf{Description} \\ 
\hline
Carbon Mass & Global, endogenous &
$\xcarbonmass_{\idstep}$,
$[\xcarbonmass^{AT}_{\idstep}, \xcarbonmass^{UP}_{\idstep}, \xcarbonmass^{LO}_{\idstep}]$ & A three dimensional vector that indicates 
the average carbon accumulation in the atmosphere, upper oceans, and lower oceans. \\  
Temperature & Global, endogenous & $\xtemperature_{\idstep}$, $[\xtemperature^{AT}_{\idstep}, \xtemperature^{LO}_{\idstep}]$ 
& A two dimensional vector that indicates the average temperature of the atmosphere and the lower ocean.  \\
Population & Regional, exogenous & $\xpopulation_{\idxi, \idstep}$ & Population and the labor in a region. \\
Technology & Regional, exogenous & $\xtechnology_{\idxi, \idstep}$ & Technology factor in the production function a region. \\
Capital & Regional, endogenous & $\xcapital_{\idxi, \idstep}$ & Total capital accumulated by a region. \\
Carbon intensity of economic activity & Regional, exogenous & $\xcarbonintensity_{\idxi, \idstep}$ & A scalar coefficient that gives the emissions resulting from economic production. \\
Balance of trade & Regional, endogenous & $\xtradebalance_{\idxi, \idstep}$ & Surplus or deficit from international trade activities. \\
Cost of mitigation efforts & Global, endogenous& $\theta_{1; \idxi, \idstep}$ & An estimate of the cost of mitigation efforts. \\
Emission due to land use & Regional, exogenous& $\xemission^{\textrm{Land}}_{\idstep}$ & Carbon emission for land use in a specific region. \\
\end{tabular}
\caption{\textbf{World-state variables.}
Global type variables correspond to the entire world, whereas regional type variables correspond to each region.
Endogenous variables are those which are affected by the agent actions whereas exogenous variables are those that are predetermined and not affected by agent actions.
Note that the values of endogenous variables can vary across steps in a predetermined manner.
\textbf{Notation: indices are separated from subscripts referring to a name by semicolons (;). For instance, the parameter $\theta_{1}$ varies in time $\idstep$ and by region $\idxi$, which is denoted as $\theta_{1; \idxi, \idstep}$.}
}
\label{tab:world_state_var}
\end{table*}

\begin{table*}[t]
\centering
\small
\begin{tabular}{p{0.28\linewidth}p{0.1\linewidth}p{0.52\linewidth}}
\textbf{Variable} & \textbf{Symbol} & \textbf{Description} \\ 
\hline
Savings rate & $\xsavings_{\idxi, \idstep}$ & The fraction of output production to be invested in capital. \\
Mitigation rate & $\xmitigation_{\idxi, \idstep}$ & The fraction of mitigation efforts by a region. \\
Import tariffs & $\xtariff_{\idxi,\idxj,\idstep}$ & The fraction of imports that are converted to tariff revenue. \\
Export limits & $\xmaxexportlimit_{\idxi, \idstep}$ & The fraction of domestic production that regions are willing to export. \\
Import bids & $\ximportbid_{\idxi,\idxj,\idstep}$ & The amount of production each region is willing to import from other regions.
\end{tabular}
\caption{\textbf{Agent-action variables.}
}
\label{tab:agent_action_var}
\end{table*}

\begin{table*}[t]
\centering
\small
\begin{tabular}{ p{0.28\linewidth} p{0.1\linewidth} p{0.52\linewidth}}
\textbf{Variable} & \textbf{Symbol} & \textbf{Description} \\ 
\hline
Initial population  & $\xpopulation_{0; \idxi}$  & The initial population for a specific region. \\
Population convergence target & $L_{a; \idxi}$ & The estimated convergence population for a specific region. \\
Population convergence rate & $ l_{g; \idxi}$& How fast the current population converges. \\
Initial capital & $\xcapital_{0; \idxi}$ & The initial capital for a specific region. \\
Initial carbon intensity & $\xcarbonintensity_{0; \idxi}$ & The initial carbon intensity for a specific region. \\
Carbon intensity parameters & $g_{\sigma; \idxi}$ and $\delta_{\sigma; \idxi}$  & The decay speed of the carbon intensity. \\
Initial technology factor & $\xtechnology_{0; \idxi}$ & The initial carbon technology factor for a specific region. \\
Technology factor parameter & $g_{\idxi, A}$ and $\delta_{\idxi, A}$  & The update pattern of the technology factor. \\
Initial land use emission & $\xemission_{L0; \idxi}$ & The initial land use emission for a specific region. \\
Land use emission parameter& $\delta_{EL; \idxi}$ & The depreciation rate for the land use emission in a specific region. \\
\end{tabular}
\caption{
\textbf{Agent-specific constants.}
}
\label{tab:agent_charact}
\end{table*}

\begin{table*}[t]
\centering
\small
\begin{tabular}{ p{0.28\linewidth} p{0.1\linewidth} p{0.52\linewidth}}
\textbf{Variable} & \textbf{Symbol} & \textbf{Description} \\ 
\hline
Capital elasticity of production & $\xcapitalelasticity$ & The contribution from capital and population to the economy. \\
Armington substitution parameter & $\lambda$ & How substitutable consumption goods from different regions are. \\
Long term welfare discount rate & $\rho$ & How much short-term welfare is weighted versus long-term welfare. \\
capital depreciation rate &  $\xcapitaldecay$ & The capital depreciation rate. \\
Backstop technology & $p_b$ & Price of a backstop technology that can remove carbon dioxide from the atmosphere. \\
Backstop technology parameter & $\delta_{pb}$ & The decay speed of the cost of backstop technology. \\
Mitigation efficiency parameter & $\theta_2$ & The efficiency loss component of mitigation \\
Domestic share parameter & $\psi^{dom}$ & The relative preference for domestic goods \\
Foreign share parameter & $\psi^{for}$ & The relative preference for foreign goods \\

\end{tabular}
\caption{
\textbf{Global constants.}
}
\label{tab:global_constants}
\end{table*}

\newpage
\section{Formal Description for Negotiations and Agreements}
\label{sec:negotiation-formal-details}

The aim of this section is to specify a working mathematical definition of a \textit{negotiation protocol} in the context of the simulator and competition.
Given the interdisciplinary nature of this work, we fully expect these definitions to grow in the future.

Consider a set of agents representing regions in RICE-N.
These agents are seeking to individually optimize their economic outcomes, but the sum of their economic outcomes depends on them collaborating toward a common climate goal.
Hence, such a setting can be understood as a variable sum, or mixed-motive, game~\cite{schelling1980strategy}.

\subsection{Agent actions}

Consider $\numAgents{}$ agents with $k$-dimensional action spaces acting from $\idstep=0$ to $\idstep=\eplen$.

Their \textbf{collective action space} at time $\idstep$ is a $k\numAgents{}$-dimensional space, where each agent's action space is a $k$-dimensional subspace of the collective action space.
The collective action space at time $\idstep$ is denoted $\ActionSpace^\idstep$, and agent $\idxi$'s action space is denoted $\ActionSpace_\idxi^\idstep$.
At each timestep $t$, an $nk$-dimensional vector is sampled from $\ActionSpace^\idstep$, with each of the $n$ agents specifying how to sample $k$ values.

Analogously, the \textbf{collective action space over time} is a $k\numAgents\eplen$-dimensional space, where each agent's action space over time is a $k\eplen$-dimensional subspace of the collective action space over time.
The collective action space over time is denoted $\ActionSpace$ and agent $\idxi$'s action space over time is denoted $\ActionSpace_\idxi$.
At the end of a run, an $nkT$-dimensional vector will have been sampled from the simulator detailing each action of each agent at each timestep.

\subsection{Agreements}
The definition of an \emph{agreement} in the context of international environmental agreements usually draws on the definition of a treaty in the 1969 Vienna convention on the law of treaties: ``an international agreement concluded between States in written form and governed by international law”~\cite{mitchell2003international}.
Notably, states participating in the agreement express a “consent to be bound”~\cite{mitchell2003international}.

Algorithmically defining agreements is challenging for numerous reasons.
In particular, legal agreements are drafted in natural language, which can not be directly interpreted by computers.
This would seemingly encourage the design of agreements through a mathematical lens to eliminate ambiguity.
However, in agreements, ambiguity is not a bug, it's a feature.
Many agreements are intentionally drafted to be generic enough to generalize to a broad set of situations~\cite{martin2020artificial}, "some of which could not have been foreseen at the time of drafting"~\cite{de2018blockchain}.

\subsection{Commitments, enforcement and incentives}

Although regions can commit to a course of action, there is no supranational body that can enforce regions to respect commitments.
Regions therefore always have the option to breach the conditions in their agreements.
Hence, regions may need to engage in strategic behavior, such as mutual enforcement or providing incentives to other regions, to maintain cooperation and adherence to agreements.
For instance, they might use trade and tariffs to reward or punish other regions.
However, understanding to what extent such behaviors are rational and strategically advantageous is an open challenge.

\subsection{Formalizing agreements}

In light of this, we provide working definitions of proposed and enacted agreements in \SimulationName{}.

A \textbf{proposed agreement} between agents $I \subseteq \{1,...,n\}$ is a set of constraints on the union of subspaces $\brckcur{ \ActionSpace_\idxi}_{\idxi \in \IndexSetI}$.



An \textbf{enacted agreement} between agents $I \subseteq \{1,...,n\}$ is a set of constraints on the sum of subspaces $\brckcur{ \ActionSpace_\idxi}_{\idxi \in \IndexSetI}$
in which each agent $\idxi \in \IndexSetI$ agrees to the constraints that apply to their subspace $\ActionSpace_\idxi$.

Naturally, an agreement enacted at time $\idstep$ can apply constraints to the entire space $\ActionSpace$, but actions that have already been drawn from $ \sum_{0}^{\idstep-1} \sum_{\idxi \in \IndexSetI} \ActionSpace_\idxi$ must be taken as given: agreements can not change the past.
For example, an agreement at time $t$ can constrain the sum total of emissions over a full run of the simulator, but it can not alter emissions from $t=0$ to $t=t-1$\footnote{In practice, reported carbon emissions are influenced by carbon accounting practices. Distinguishing between ground truth emissions, regions' private climate-economic information and what they make publicly available within the simulator is an interesting area of exploration.}.

A key distinction between \emph{proposed} agreements and \emph{enacted} agreements is that, while proposed agreements can be generated in any manner, enacted agreements must be voluntarily accepted by the agents constrained by the agreement, expressing the "consent to be bound" aspect of the Vienna convention on the law of treaties~\cite{mitchell2003international}.
In other words, a proposed agreement cannot be enacted (become an enacted agreement) without the consent of all implicated agents. 
If a proposed agreement is not enacted, then its constraints are not applied in the simulation.

%
Currently, an enacted agreement only applies changes to $\ActionSpace^\idstep$.
In the next time period, new proposed agreements are submitted and evaluated, which enables regions to not follow through on their previous agreements.

\subsection{Agreements through negotiation protocols}
A \textbf{negotiation protocol} is an algorithmic procedure to propose agreements.
The goal of such a protocol is to propose agreements that agents will enact and that lead to optimal climate-economic outcomes by balancing collective climate and individual economic objectives.

Negotiation protocols currently consist of two (potentially iterative) phases: 1) generating proposals, and 2) evaluating proposals.

\paragraph{Generating proposals.} 
Agreements can be generated in any way that participants see fit.
Such agreements can be proposed 1) \emph{internally}, which is to say that agents participating in the negotiation protocol can send messages to other agents in which they propose agreements, or
2) \emph{externally}, where an external body (represented within the negotiation protocol itself) proposes an agreement to a subset of participating agents.

\paragraph{Evaluating proposals.} 
Agents participating in the simulation must evaluate proposals that concern them to determine whether they wish to accept them or refuse them.
Importantly, a proposed agreement that is refused by an agent $\idxi$ must not apply constraints to the action space $\ActionSpace_\idxi$ of the agent that refused it.
However, an agreement proposed internally by a participating agent is de facto considered accepted by that agent.
For example, an agent might partition their own action space and propose an agreement to another agent while guaranteeing that they will only choose actions within a certain partition based on the other agent's acceptance or refusal.

\paragraph{Repeating the process.}
Generation and evaluation can iterate in a negotiation protocol.
For example, an agent $\idxi$ could initially propose an agreement to an agent $\idxj$.
Agent $\idxj$ can refuse this agreement, and propose an alternative agreement back to $\idxi$, or to another agent $\idxk$.

Generating a proposed agreement that agents will enact can be seen as a double constraint satisfaction problem: the goal of a negotiation protocol is to identify a set of constraints in the action space that are accepted by the participating agents.
Agents map proposed agreements implicating them to either 0 or 1 depending on acceptance or refusal.
The agreement is itself a set of constraints on the action space.
Therefore, generating a proposed agreement is equivalent to generating a set of constraints on the action space that is mapped to the intersection of the acceptance space for all implicated agents.

This implies a curse of dimensionality when it comes to generating agreements: since an agreement fails unless all concerned agents accept, the more agents an agreement is proposed to, the less likely it is that the agreement is enacted.





\subsection{Communication}

Within \SimulationName{}, agents may communicate by exchanging messages, which consist of numerical values that do not directly affect the climate-economic dynamics 
of the simulator.
However, such messages might influence subsequent actions that directly impact the dynamics.

The goal of the competition can therefore be broadly understood as attempting to structure communication within the simulation in a way that, e.g., agents might share information about goals or priorities, their willingness to cooperate, with the end goal of aligning incentives and maximizing their individual objectives, and ultimately global climate and economic goals.

Modeling and learning how to communicate in multi-agent RL is its own subfield.
We refer readers to ~\cite{zhu2022survey} for a recent review of the topic, and~\cite{foerster2018deep} for an in-depth technical introduction.


\end{document}